\documentclass[10pt,twocolumn,letterpaper]{article}
\usepackage[pagenumbers]{cvpr} %

\usepackage[dvipsnames,svgnames]{xcolor}
\newcommand{\red}[1]{{\color{red}#1}}
\definecolor{darkgreen}{rgb}{0.0, 0.75, 0.0} %

\newcommand{\green}[1]{{\color{darkgreen}#1}}
\newcommand{\blue}[1]{\textcolor{blue}{#1}}

\definecolor{darkyellow}{rgb}{0.75, 0.75, 0.0} %
\newcommand{\yellow}[1]{\textcolor{darkyellow}{#1}}

\newcommand{\myparagraph}[1]{\smallskip\noindent\textbf{#1}}

\usepackage{colortbl}
\newcommand{\myrowcolour}{\rowcolor[gray]{0.925}}

\newcommand{\aftertab}{\vspace{-1.75em}}
\newcommand{\afterfig}{\vspace{-1em}}
\newcommand{\aroundeqn}{\vspace{0em}}
\newcommand{\aftertabcaption}{\vspace{-0.75em}}
\newcommand{\beforetfigcaption}{\vspace{-0.5em}}
\makeatletter
\DeclareRobustCommand\onedot{\futurelet\@let@token\@onedot}
\def\@onedot{\ifx\@let@token.\else.\null\fi\xspace}

\def\eg{\emph{e.g}\onedot} 
\def\ie{\emph{i.e}\onedot} 
 
\def\etc{\emph{etc}\onedot}

\def\etal{\emph{et al}\onedot}
\makeatother

\newcommand{\term}{geometry}
\newif\ifdrafting
\draftingtrue %
\draftingfalse %
\ifdrafting
    \newcommand{\todo}[1]{{\leavevmode\color[rgb]{1,0,0}[TODO: #1]}}
    \newcommand{\ds}[1]{{\leavevmode\color[rgb]{0.8,0,0.8}[Deqing: #1]}}
    \newcommand{\vj}[1]{{\leavevmode\color[rgb]{0.18,0.64,0.37}[Varun: #1]}}
    \newcommand{\cih}[1]{{\leavevmode\color[rgb]{0,0.5,0}[Charles: #1]}}
    \newcommand{\jh}[1]{{\leavevmode\color[rgb]{0.8, 0.2, 0}[Junhwa: #1]}}
    \newcommand{\jz}[1]{{\leavevmode\color[rgb]{0,0.6,0.8}[Junyi: #1]}}
    \newcommand{\mh}[1]{{\leavevmode\color[rgb]{0,0.8,0.8}[MHY: #1]}}
    
\else
    \newcommand{\todo}[1]{}
    \newcommand{\ds}[1]{}
    \newcommand{\vj}[1]{}
    \newcommand{\cih}[1]{}
    \newcommand{\jh}[1]{}
    \newcommand{\jz}[1]{}
    \newcommand{\mh}[1]{}
    
\fi
\usepackage{multirow}
\usepackage{array}
\usepackage{subcaption}
\usepackage[accsupp]{axessibility}
\newcolumntype{C}[1]{>{\centering\arraybackslash}p{#1}}
\definecolor{cvprblue}{rgb}{0.21,0.49,0.74}
\usepackage[pagebackref,breaklinks,colorlinks,citecolor=cvprblue,linkcolor=cvprblue]{hyperref}

\def\paperID{475} %
\def\confName{CVPR}
\def\confYear{2024}

\newcommand{\ignore}[1]{}

\title{Telling Left from Right:
Identifying Geometry-Aware Semantic Correspondence}

\author{
Junyi Zhang$^{\dagger}$ \qquad Charles Herrmann$^{\ddagger}$ \qquad Junhwa Hur$^{\ddagger}$ \qquad Eric Chen$^{\mathsection}$\\ 
\qquad Varun Jampani$^{\mathparagraph}$ \qquad Deqing Sun$^\ddagger\ $\thanks{Equal contribution.} \qquad Ming-Hsuan Yang$^{\ddagger,\mathsection\ *}$ \qquad \\
\vspace{-0.75em}\\
$^{\dagger}$Shanghai Jiao Tong University \ \ \ $^{\ddagger}$Google Research \ \ \ $^{\mathsection}$UIUC \ \ \ $^{\mathparagraph}$Stability AI \ \ \ $^{\mathsection}$UC Merced
}

\begin{document}
\maketitle

\begin{abstract}
While pre-trained large-scale vision models have shown significant promise for semantic correspondence, their features often struggle to grasp the geometry and orientation of instances.
This paper identifies the importance of being geometry-aware for semantic correspondence and reveals a limitation of the features of current foundation models under simple post-processing.
We show that incorporating this information can markedly enhance semantic correspondence performance with simple but effective solutions in both zero-shot and supervised settings. 
We also construct a new challenging benchmark for semantic correspondence built from an existing animal pose estimation dataset, for both pre-training validating models. 
Our method achieves a PCK@0.10 score of \textbf{65.4} (zero-shot) and \textbf{85.6} (supervised) on the challenging SPair-71k dataset, surpassing the state of the art by 5.5p and 11.0p absolute gains, respectively.
Our code and datasets are publicly available at: \url{https://telling-left-from-right.github.io}.
\end{abstract}
\vspace{-1em}
\section{Introduction}
\label{sec:intro}

\begin{figure}[t]
  \centering
  \begin{subfigure}{0.975\linewidth}
    \includegraphics[width=\linewidth]{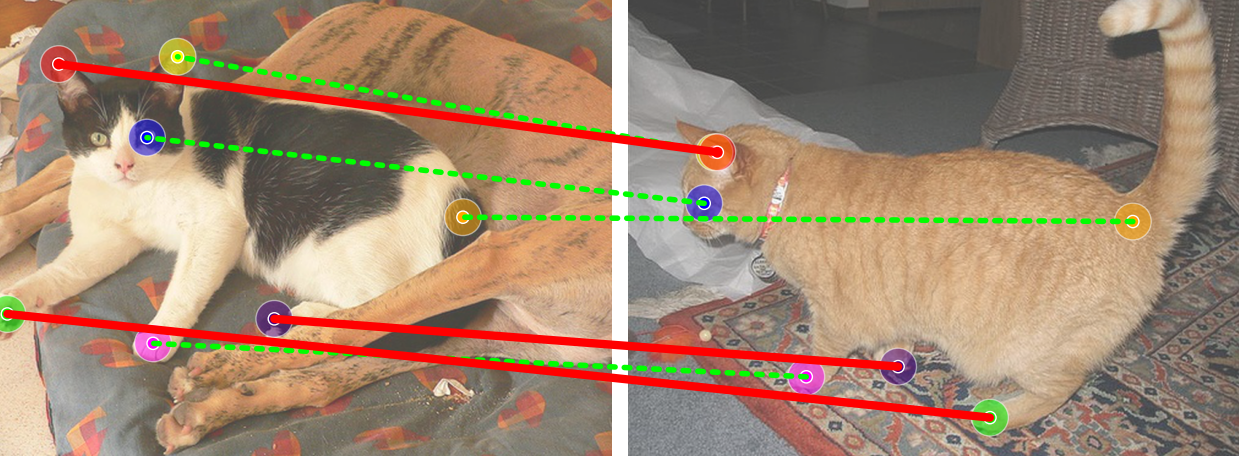}
    \caption{The state-of-the-art method~\cite{zhang2023tale} fails at matching keypoints with geometric ambiguity, or, ``telling left from right" (\red{red solid lines}). 
    }
    \label{fig:1.1a}
  \end{subfigure}
  \begin{subfigure}{0.975\linewidth}
    \includegraphics[width=1.0\linewidth]{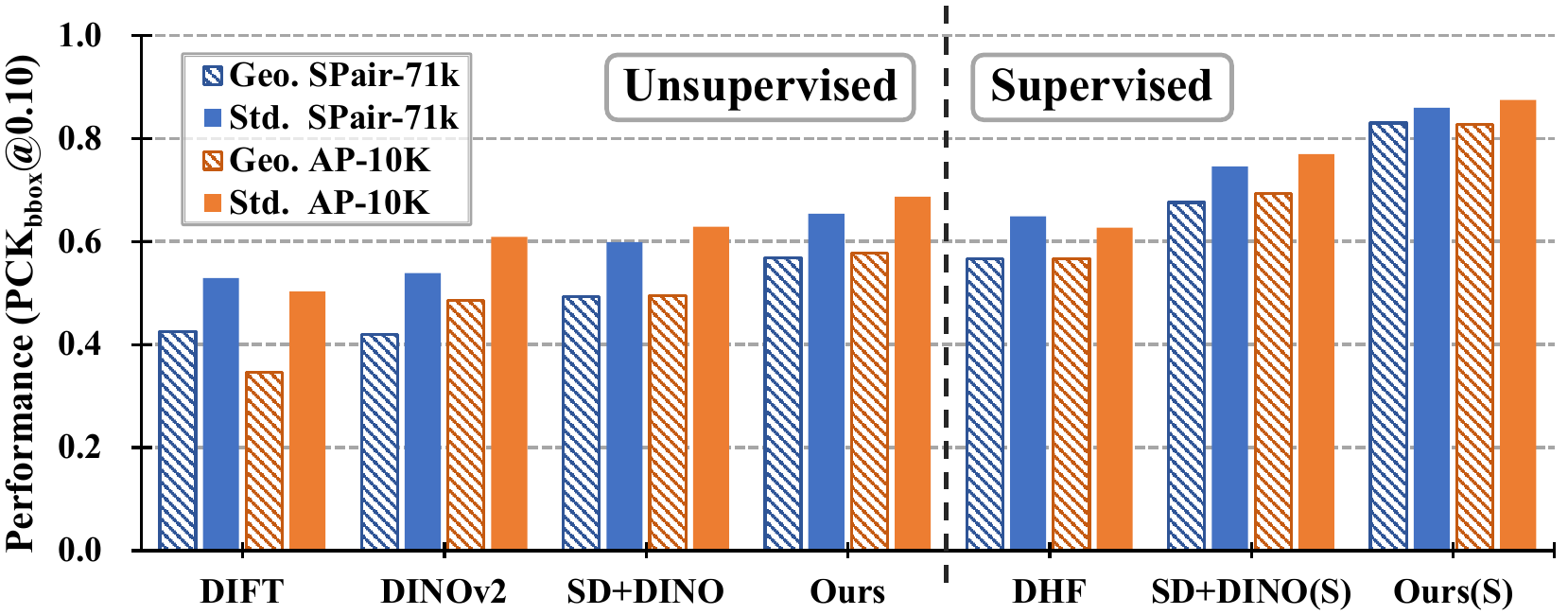}
    \caption{\textbf{The performance gap between geometry-aware set (Geo.) and standard set (Std.) of state-of-the-art methods.} %
    The geometry-aware set accounts for \textbf{{59.6\%}} and \textbf{{45.7\%}} of the total keypoint pairs on SPair-71k~\cite{min2019spair} and AP-10K~\cite{yu2021ap}, respectively.}
   \label{fig:1.2performance}
  \end{subfigure}
  \beforetfigcaption
  \caption{Illustration of geometry-aware correspondence.}
  \label{fig:1.1illustration}
  \vspace{-0.25em}
  \afterfig
\end{figure}

Since the advent of high fidelity text-to-image (T2I) generative models~\cite{saharia2022photorealistic, rombach2022high} and large vision foundation models~\cite{oquab2023dinov2}, there has been significant interest in understanding both what these models are learning and what they are not. Numerous works show that these models have powerful feature embeddings that can be used for many computer vision tasks including depth estimation~\cite{zhao2023unleashing,oquab2023dinov2}, semantic segmentation \cite{tian2023diffuse, xu2023open}, and semantic correspondences \cite{amir2021deep,hung2019scops, ofri2023neural, gupta2023asic, zhang2023tale,luo2023diffusion,hedlin2023unsupervised}. While many works have shown their strengths, less analysis has been done on their weaknesses; in particular, what do these features struggle with?

We propose using %
semantic correspondence as a promising test bed. Semantic correspondence, the establishment of pixel-level matches between two images with semantically similar objects, is an important Computer Vision problem with a variety of downstream applications, \eg, image editing~\cite{gupta2023asic,ofri2023neural,zhang2023tale,mou2023dragondiffusion} and style transfer~\cite{lee2020reference, geyer2023tokenflow}.\ignore{, instance segmentation~\cite{hong2022cost,chen2020show,lan2021discobox}, and part segmentation~\cite{amir2021deep,sun2023going}.} It also has many difficult challenges, \eg, from large intra-class variations to different backgrounds, lighting, or viewpoints.

Despite these challenges, the large foundation model features currently achieve state-of-the-art performance~\cite{zhang2023tale}. However, a closer examination shows that this performance is inconsistent across all challenges. In particular, we find that these foundation model's features significantly underperform on ``geometry-aware''\footnote{We use the term geometry in the loose sense and do not refer to 3D geometric properties, such as shape and surface normal.} semantic correspondences: correspondences which share semantic properties but have different relations to the overall geometry of the object, \eg, the ``left'' paw vs. the ``right'' paw as shown in \cref{fig:1.1a}.
Motivated by this, we conduct an in-depth analysis of these correspondences (\cref{fig:1.2performance}). We find that surprisingly, such cases account for a significant portion of the benchmark datasets (nearly 60\% in SPair71k), and state-of-the-art methods with the deep features perform considerably worse on this challenging subset (up to 30\% worse, \cref{fig:1.2performance}).

Are these problems an innate failing of these features, or can they be alleviated through better post-processing? Based on the above observations, we propose several methods that resolve the geometric ambiguity during matching.
First, we introduce a test-time viewpoint alignment strategy that approximately aligns viewpoints of instances to make the problem easier. 
Then we train a lightweight post-processing module that improves geometric awareness of features from visual foundation models \cite{oquab2023dinov2,rombach2022high}, by using a soft-argmax based dense training objective with given annotated sparse keypoints. 
We further introduce a pose-variant augmentation strategy as well as a window soft-argmax module.
These not only significantly improve performance on standard benchmarks by 15\% while costing only $0.32\%$ of extra runtime.

For more advanced analysis, we create a new benchmark dataset using existing annotations from the AP-10K~\cite{yu2021ap} animal pose estimation dataset.
Compared to the largest existing benchmark~\cite{min2019spair}, our new benchmark dataset includes 5 times more training pairs and also evaluates cross-species and cross-families semantic correspondence in addition to intra-species correspondence.
We also demonstrate that this benchmark can serve as a valuable pre-training resource for improving geometry-aware semantic correspondence.

To summarize, we make the following contributions:
\begin{itemize}
    \item We identify the problem of geometry-aware semantic correspondence and show that pre-trained features of foundation models (SD~\cite{rombach2022high} and DINOv2~\cite{oquab2023dinov2}) struggle with geometric information. %
    \item We propose to improve geometric awareness of the features in both unsupervised and supervised manners.
    \item We introduce a large-scale and challenging benchmark, AP-10K, for both training and evaluation. %
    \item Our method boosts the overall performance on multiple benchmark datasets, especially on the \term-aware correspondence subset. %
    It achieves an 85.6 PCK@0.10 score on SPair-71k, outperforming the state-of-the-art method by more than 15\%.
\end{itemize}

\section{Related Work}
\label{sec:related}

\vspace{-0.5em}
\myparagraph{Semantic correspondence.}
Conventional approaches to semantic correspondence estimation follow a common pipeline that consists of i) feature extraction~\cite{dalal2005histograms,lowe1999object,simonyan2014very,long2014convnets,he2016deep,aberman2018neural}), ii) cost volume computation \cite{hong2022cost,cho2021cats,lee2019sfnet,min2020learning}, and iii) matching field regression \cite{truong2020glu,kim2019semantic,truong2021learning,truong2020gocor,truong2021warp}.
To handle challenging intra-class variations between images, previous work have presented various approaches such as matching uniqueness prior~\cite{liu2020semantic}, parameterized spatial prior~\cite{seo2018attentive,rocco2018end,rocco2017convolutional,kim2018recurrent,huang2022learning, yang2017object}, or end-to-end regression \cite{lee2021patchmatch, cho2021cats, cho2022cats++, huang2022learning,truong2020glu}. 
A few previous work have also explored semantic correspondence under unsupervised~\cite{shtedritski2023learning, thewlis2017unsupervised, gupta2023asic, ofri2023neural} and weakly supervised~\cite{peebles2022gan, huang2023weakly, truong2022probabilistic} setting, by densely image aligning~\cite{thewlis2017unsupervised,  gupta2023asic, ofri2023neural, peebles2022gan} or automatic label generation~\cite{shtedritski2023learning, huang2023weakly}.
However, due to the limited capacity of their features or the usage of strong spatial prior, they still exhibit difficulties handling challenging intra-class variations such as large pose changes or non-rigid deformation.

\begin{figure}[t]
  \centering
   \includegraphics[width=1.0\linewidth]{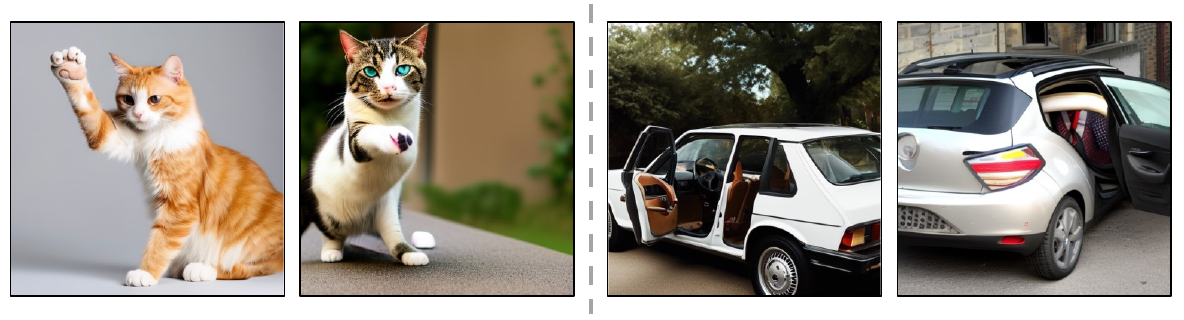}
   \vspace{-1.25em}
   \beforetfigcaption
   \caption{Generated samples from SD-2-1 with the prompt (left) ``A cat holding up its \textit{left front paw}" and (right) ``A car with the \textit{right front door} open". %
   SD has difficulty generating images that require understanding the intrinsic geometry of instances. 
   }
   \label{fig:2related}
   \afterfig
   \vspace{-0.25em}
\end{figure}

Recently, visual foundation models (\eg DINO \cite{caron2021emerging, oquab2023dinov2} and SD \cite{rombach2022high}) demonstrate that their pretrained features, learned by self-supervised learning or generative tasks \cite{luo2023diffusion, zhang2023tale, hedlin2023unsupervised, amir2021deep, tang2023dift}, can serve as powerful descriptors for semantic matching by surpassing prior arts specifically designed for semantic matching.
Yet, we reveal that such models still show limitations \cite{gupta2023asic} in comprehending the intrinsic geometry of instances (\eg,~\cref{fig:2related}) and formally investigate this issue, termed ``geometry-aware" semantic correspondence.

\myparagraph{Benchmark datasets.}
Recent advances in semantic correspondence have continuously revealed limitations of existing benchmark datasets. 
For example, widely used datasets (PF-Pascal~\cite{ham2016proposal}, PF-Willow~\cite{ham2017proposal}, CUB-200-2011~\cite{wah2011caltech}, and TSS~\cite{taniai2016joint}) provide image pairs with only limited viewpoints or pose variations, making it hard to evaluate methods on handling large object viewpoint changes.
The CUB dataset~\cite{wah2011caltech} provides images of a single object class, bird, only.
SPair-71k~\cite{min2019spair} introduces a more challenging benchmark dataset that consists of 1,800 images across 18 object categories with substantial intra-class variations.
Recently, Ayg{\"u}n and Mac Aodha~\cite{aygun2022demystifying} leveraged an animal pose dataset, Awa-Pose~\cite{banik2021novel}, to create 10k image pairs for evaluating the inter-class semantic correspondence.
While existing methods~\cite{zhang2023tale, luo2023diffusion, cho2022cats++, aygun2022demystifying} have low performance on the SPair-71k and Awa-Pose, these benchmarks are still small-scale.
To address these shortcomings, we introduce a new, large-scale, and challenging benchmark using the animal pose estimation dataset, AP-10K~\cite{yu2021ap}. 
This new benchmark further facilitates comprehensive evaluations of geometric awareness and provides annotations for training models.

\section{Geometric Awareness of Deep Features}
\label{sec:geo-aware}

In this section, we first provide the clear problem definition of ``geometry-aware semantic correspondence" as challenging cases of semantic correspondence, which requires an understanding of relations of similar semantic parts.
Then we provide comprehensive analyses on the performance of pretrained features of foundation models on the problem and what geometric information those features possess.

\subsection{Geometry-Aware Semantic Correspondence}

\begin{figure}[t]
  \centering
   \includegraphics[width=0.9\linewidth]{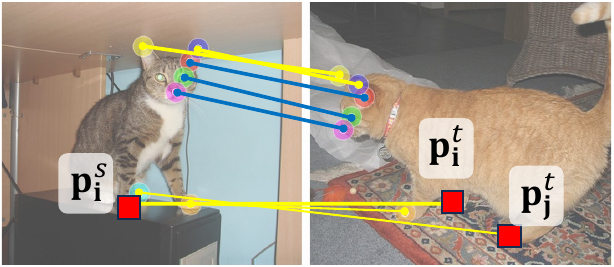}
    \beforetfigcaption
   \caption{Annotations of geometry-aware semantic correspondence (\yellow{yellow}) and standard semantic correspondence (\blue{blue}). 
   }
   \label{fig:3.1illustration}
   \afterfig
\end{figure}

We define geometry-aware semantic correspondence as a challenging case of semantic correspondence, where there exist geometry-ambiguous matching cases, and thus it requires an understanding of instances' orientations or geometry.
\cref{fig:1.1a} illustrates the exemplar cases that require proper understanding of both semantic parts (\ie paws) and their associations (\ie left paw or right paw) with the orientation of instances.

As a formal definition, for each instance category, we first cluster keypoints into subgroups $\mathcal{G}_{parts}$ by their semantic parts.
Each subgroup $\mathcal{G}_{parts}$ consists of a set of keypoints $\mathbf{p}_{({parts},~{index})}$ that fall into the same subgroup but position in different part locations according to their orientations.
For the \emph{cat} category as an example, the subgroups are $\mathcal{G}_{parts}$ with ${parts}$ = \{ears, paws, ... \}, and $\mathcal{G}_\text{paws}$ = \{$\mathbf{p}_{(\text{paws},~\text{front left})}$, $\mathbf{p}_{(\text{paws},~\text{front right})}$, $\mathbf{p}_{(\text{paws},~\text{rear left})}$, $\mathbf{p}_{(\text{paws},~\text{rear right})}$\}.

Then, give a source $\mathbf{I}^s$ and a target image $\mathbf{I}^t$ that contains the same/similar instance category with their keypoint correspondence annotations, the correspondence $\langle \mathbf{p}_\mathbf{i}^{s}, \mathbf{p}_\mathbf{i}^{t} \rangle$ is considered as a ``geometry-aware'' correspondence if the two conditions are met.
First, two keypoints $\mathbf{p}_\mathbf{i}^s$ and $\mathbf{p}_\mathbf{i}^t$ belong to the same subgroup, $\mathbf{p}_\mathbf{i}^s \in \mathcal{G}_{part}^s$ and $\mathbf{p}_\mathbf{i}^t \in \mathcal{G}_{part}^t$. Second, there are other visible keypoint(s) belonging to same subgroup in the target image, $\ \exists \ \mathbf{j} \neq \mathbf{i} \text{ s.t. }  \mathbf{p}_\mathbf{j}^t \in \mathcal{G}_{part}^t$.
As illustrated in \cref{fig:3.1illustration}, the front right paw ($\mathbf{p}^s_\mathbf{i}$) of the cat in the source image has several semantically similar correspondences, such as ($\mathbf{p}^t_\mathbf{j}$) and ($\mathbf{p}^t_\mathbf{i}$), which requires proper understanding of geometry to find the correct match.

\subsection{Evaluation on the Geometry-aware Subset}

We evaluate the state-of-the-art methods on geometry-aware semantic correspondence to see if their features are geometry-aware and how well they perform on this challenging task.
From the challenging SPair-71k~\cite{min2019spair} datasets, we first cluster keypoint subgroups $\mathcal{G}_{parts}$ for each category and gather geometry-aware correspondence cases as the ``geometry-aware subset''.
Surprisingly such cases account for a substantial portion, $82.4\%$ of total image pairs and $59.6\%$ of matching keypoints, of the dataset. (Please refer to Supp.~\ref{sec:geo-aware-details} for more details.) %

\begin{figure}[t]
  \centering
   \includegraphics[width=1.0\linewidth]{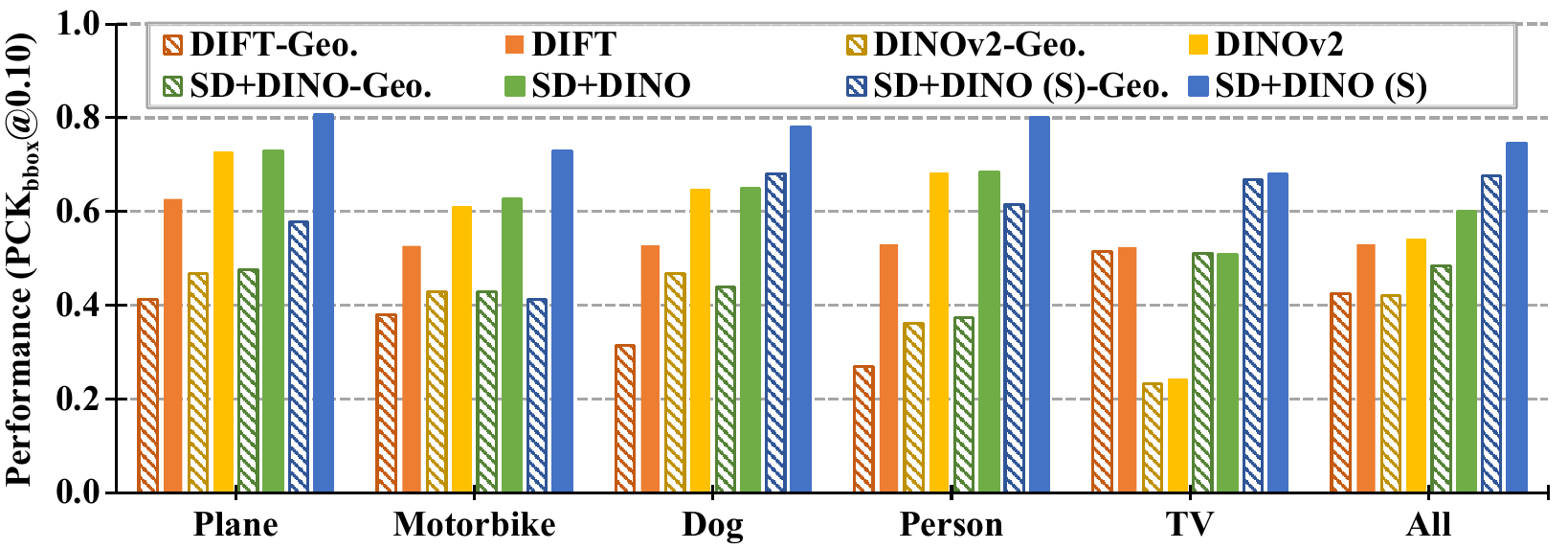}
   \vspace{-1.25em}
   \beforetfigcaption
   \caption{\textbf{Per-category evaluation of state-of-the-art methods on SPair-71k geometry-aware subset (Geo.) and standard set.} While the geometry-aware subset accounts for 60\% of the total matching keypoints, we observe a substantial performance gap between the two sets for all the methods.
   }
   \label{fig:3.2per-category}
   \afterfig
\end{figure}

\cref{fig:3.2per-category} shows the performance of the state-of-the-arts on the subset (in both zero-shot and supervised (S)).
For all methods, there exists a substantial performance gap between the geometry-aware subset and the standard set, around 20\% for zero-shot methods and still 10\% for supervised methods.
This reveals the weakness of current methods in matching keypoints where the geometry ambiguity is involved and the limitation on geometric awareness.

\subsection{Sensitivity to Pose Variation}
\label{sec:sensitive_to_pose}

\begin{figure}[t]
  \centering
   \includegraphics[width=0.975\linewidth]{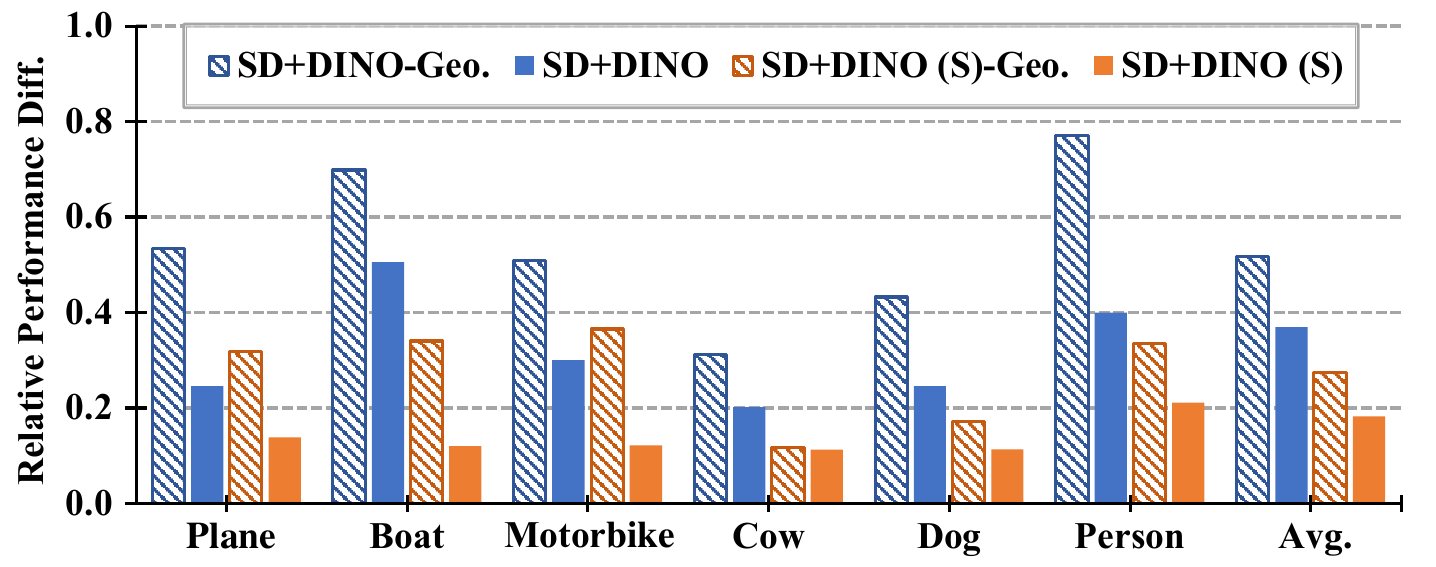}
   \vspace{-0.25em}
   \beforetfigcaption
   \caption{\textbf{Evaluation of the sensitivity to pose variations.} 
   The y-axis shows the normalized difference between the best and the worst performance among 5 different azimuth-variation subsets.
   We report the results of the unsupervised and supervised methods on both the geometry-aware (Geo.) and standard set.
   The larger the value, the more sensitive the performance is to pose variation.
   }
   \label{fig:3.3per-category-az}
   \afterfig
\end{figure}

For certain categories, however, where the pose variation of the pair images is small (\eg, potted plant and TV in SPair-71k), performance gaps on both the standard and geometry-aware sets are nearly marginal.
This suggests that the pose variation is one of the key factors that affect the accuracy of geometry-aware correspondence.
To delve deeper into it, we divide image pairs in SPair-71k into 5 subsets, based on their annotated azimuth differences, ranging from 0 (identical poses) to 4 (completely opposing directions).
For each category, we then again evaluate the performance on these 5 subsets, $\mathcal{A}=\{\mathbf{a_0}, \mathbf{a_1}, \dots, \mathbf{a_4}\}$ and define the normalized relative difference, $\textbf{d} \!=\!\frac{\max(\mathcal{A}) - \min(\mathcal{A})}{\max(\mathcal{A})}$, which measures the sensitivity to pose variations.
As shown in \cref{fig:3.3per-category-az}, 
the performance on the geometry-aware subset is more sensitive to the pose variation than the standard set across all categories, indicating that the pose variation affects the performance on geometry-aware semantic correspondence.

\subsection{Global Pose Awareness of Deep Features}
\label{subsec:global_pose_awareness}
\begin{figure}[t]
  \centering
   \includegraphics[width=1\linewidth]{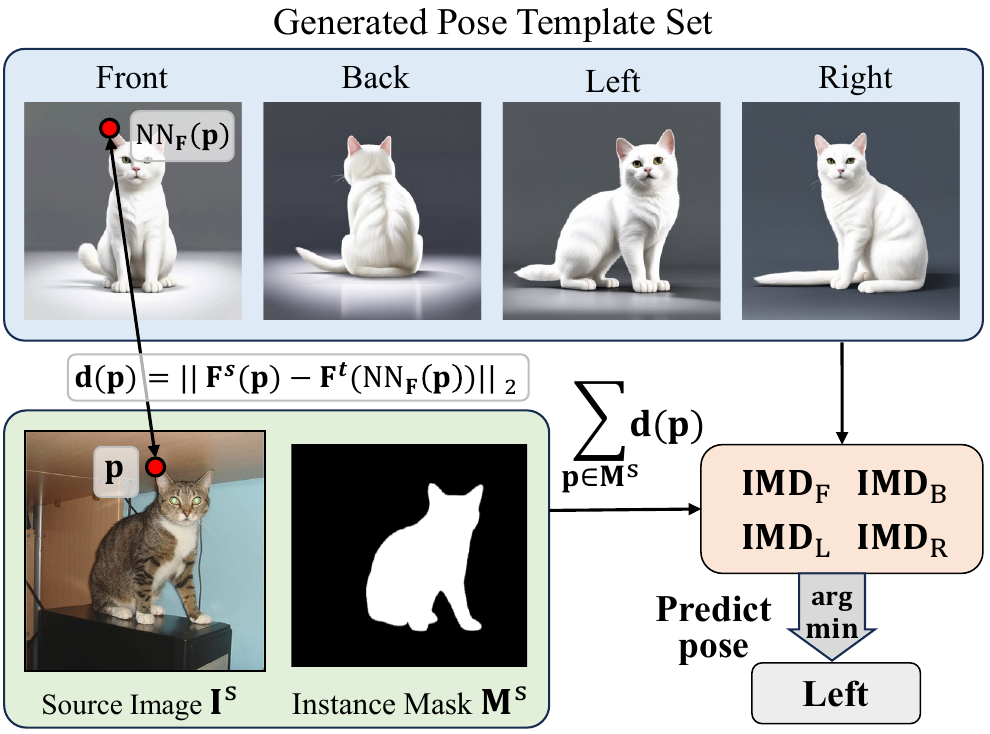}
   \vspace{-1.5em}
   \caption{\textbf{Rough pose prediction with feature distance.} By computing the instance matching distance (IMD) of the source image to the generated pose templates, we can utilize the feature maps to predict rough pose and evaluate the global pose awareness of current deep features. We only show one template set for brevity.
   }
   \label{fig:3.5imd}
   \vspace{-0.25em}
   \afterfig
\end{figure}
We further analyze if deep features are aware of high-level pose (or viewpoint) information of an instance in an image.
We explore this pose awareness by a template-matching approach in the feature space.

\myparagraph{Instance matching distance (IMD).}  
We introduce this metric to examine pose prediction accuracy.
Given a source $\mathbf{I}^s$ and target image $\mathbf{I}^t$, their normalized feature maps $\mathbf{F}^s$ and $\mathbf{F}^t$, and a source instance mask $\mathbf{M}^s$, we define the IMD metric as:
\aroundeqn
\begin{equation}
\operatorname{IMD}(\mathbf{I}^s, \mathbf{I}^t, \mathbf{M}^s) \!=\!\sum_{\mathbf{p} \in \mathbf{M}^s} \| \mathbf{F}^s(\mathbf{p})\!-\! \text{NN}(\mathbf{F}^s(\mathbf{p}), \mathbf{F}^t) \|_2, 
\label{eq:idm}
\end{equation}
\aroundeqn
where $\mathbf{p}$ denotes a pixel within the source instance mask, $\mathbf{F}^s(\mathbf{p})$ is the feature vector at $\mathbf{p}$, and $\text{NN}(\mathbf{F}^s(\mathbf{p}), \mathbf{F}^t)$ represents the nearest-neighboring feature vector in the target feature map. 
IMD measures the similarity of two images via the average feature distance of corresponding pixels.

\myparagraph{Pose prediction via IMD.}
With the IMD metric, we can evaluate the global pose awareness of features from existing methods via pose prediction.
We start by generating multiple pose template sets (in~\cref{fig:3.5imd}). We then compute the IMD between the input and template images for each set and predict the pose whose IMD is the smallest. A collective vote across all sets determines the final pose estimate.

We manually annotated 100 cat images from SPair-71k with pose labels \{left, right, front, and back\} and evaluate the pose prediction performance of the following deep features: DINOv2~\cite{oquab2023dinov2}, SD~\cite{rombach2022high}, and fused SD+DINO~\cite{zhang2023tale}.
Due to some ambiguous cases for annotations, we also report the performance of binary classification into \{left, right\} or \{front, back\}.
As in~\cref{tab:pose_prediction}, DINOv2 struggles with left/right (L/R) distinction~\cite{gupta2023asic} but excels in front/back (F/B) prediction; 
SD performs well in both distinguishing L/R and F/B; 
and SD+DINO surpass both in all cases, achieving near-perfect results. 
This suggests that the deep features are aware of global pose information.

\begin{table}
\centering
\small
\caption{\textbf{Zero-shot rough pose prediction result with IDM (\cref{eq:idm}).} We report the accuracy of predicting left or right (L/R), front or back (F/B), the former two cases (L/R or F/B), and one of the four directions (L/R/F/B). }
\vspace{-0.75em}
\label{tab:pose_prediction}
\resizebox{0.4\textwidth}{!}{
\begin{tabular}{lcccc}
\toprule
 Feature       & L/R   & F/B   & L/R or F/B & L/R/F/B\\
\midrule
DINOv2  & 63.8  & 100.0 & 75.0 & 51.0    \\
SD      & 95.7  & 96.8  & 96.0  & 78.0    \\
SD+DINO & 98.6 & 100.0 &  99.0 & 84.0     \\
\bottomrule
\end{tabular}
}
\vspace{0.75em}
\aftertab
\end{table}
\section{Improving Geo-Aware Correspondence}
\label{sec:supervised}

\begin{figure}[t]
  \centering
   \includegraphics[width=0.925\linewidth]{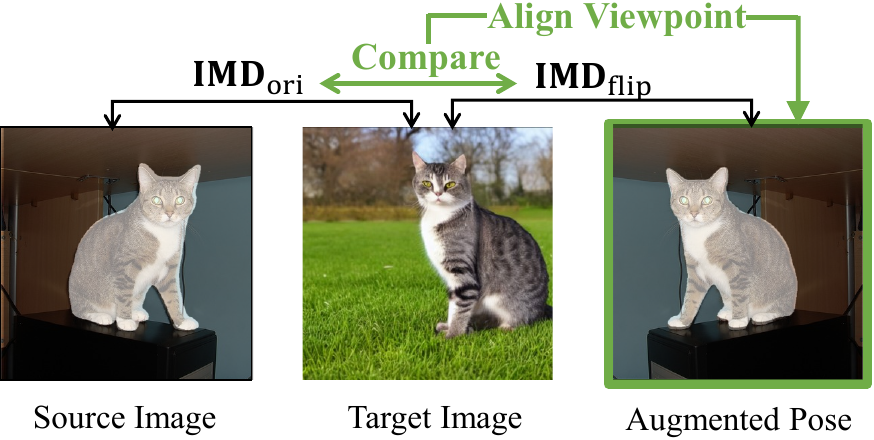}
   \beforetfigcaption
   \caption{\textbf{Adaptive pose alignment.} By comparing the matching distance between the target image and augments of the source image, we can reduce the pose variation of pair images at test time for better correspondence.}
   \label{fig:4.5pose}
   \afterfig
   \vspace{-0.5em}
\end{figure}

We propose several techniques that improve geometric awareness during matching, in both zero-shot and supervised settings.
We first introduce an adaptive pose alignment strategy that runs at test time without any training involved.
Then, we further introduce a post-processing module with various training strategies that can improve the geometry awareness of deep features.

\begin{figure*}[t]
  \centering
   \includegraphics[width=1\linewidth]{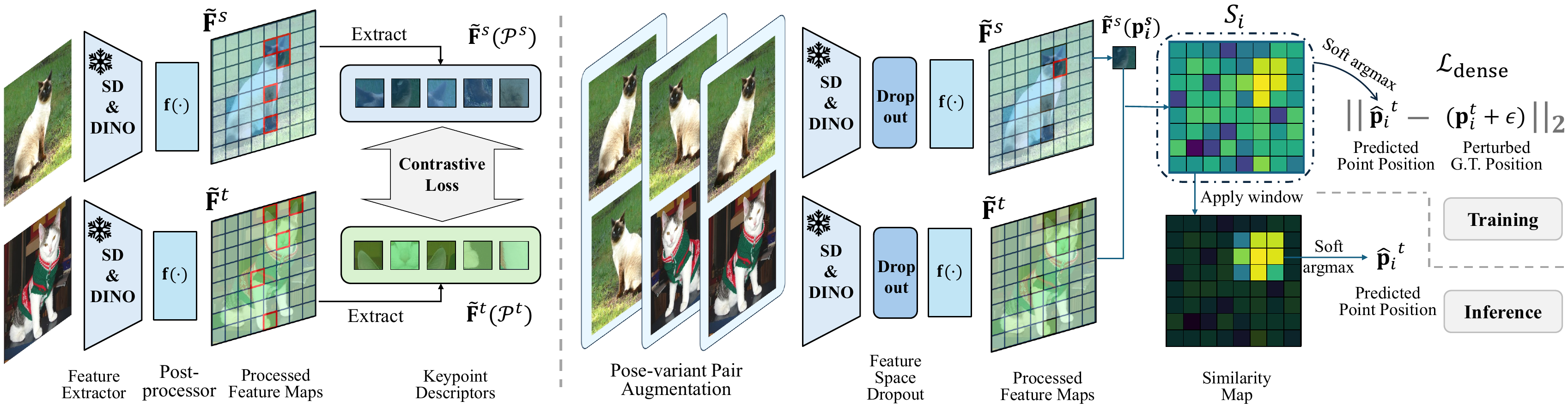}
   \vspace{-1.5em}
   \beforetfigcaption
   \caption{
   (Left) previous supervised methods~\cite{luo2023diffusion,zhang2023tale} with a sparse training objective. (Right) an overview of our supervised method. Only the lightweight post-processor is updated during training. 
   Both the pair augmentation and feature space Dropout are for training only.
   }
   \label{fig:5supervised}
   \afterfig
   \vspace{-0.75em}
\end{figure*}

\begin{figure}[t]
  \centering
   \includegraphics[width=0.90\linewidth]{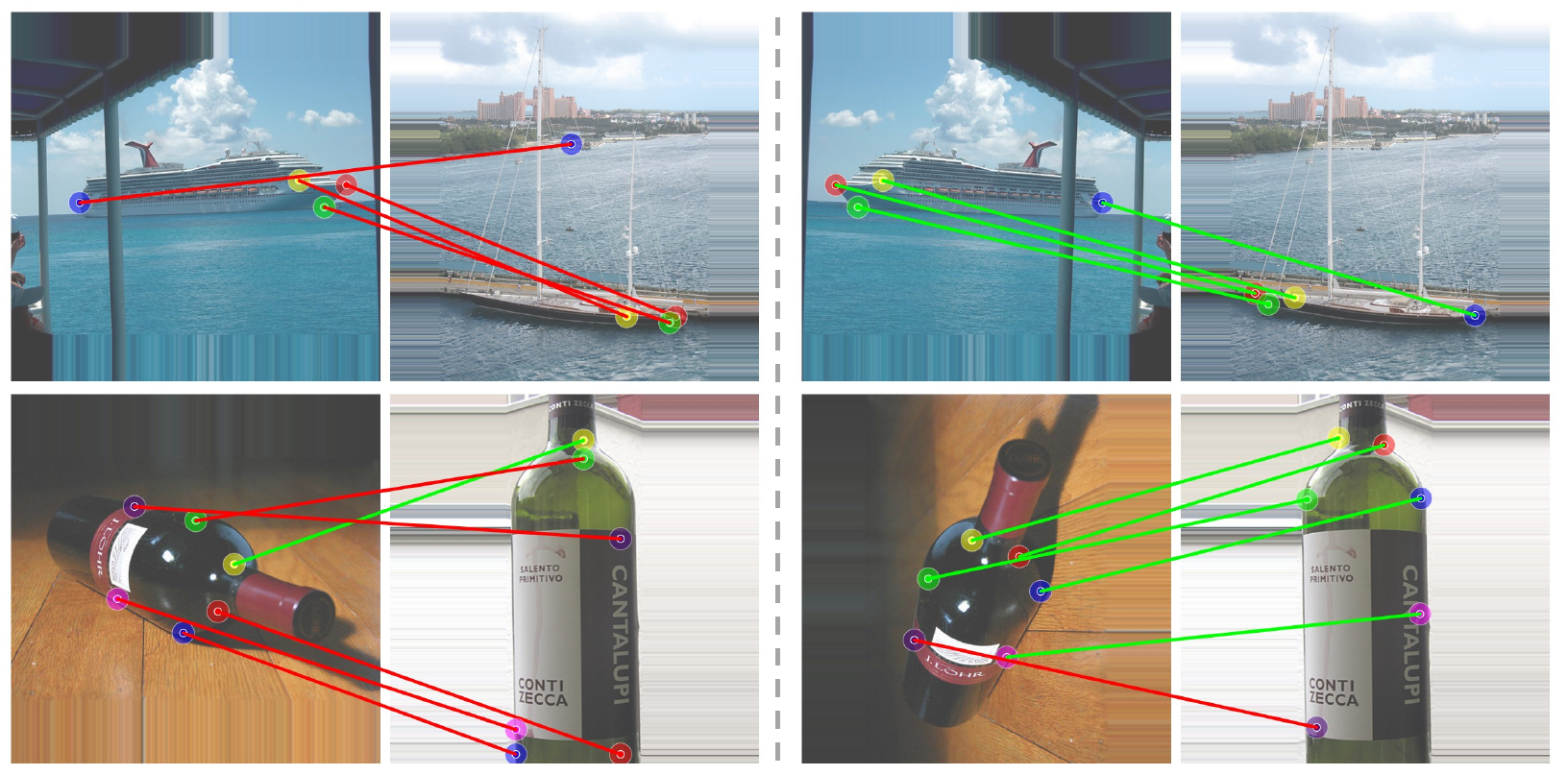}
   \vspace{-0.5em}
   \beforetfigcaption
   \caption{(Left) original image pairs. (Right) image pairs with the test-time aligned pose. The reduced pose variation improves the correspondence accuracy.}
   \label{fig:4zeroshot_exp}
   \vspace{-0.5em}
   \afterfig
\end{figure}

\subsection{Test-time Adaptive Pose Alignment}
\label{sec:pose_alignment}
In~\cref{sec:sensitive_to_pose}, we find that pose variations can largely affect the performance of geometry-aware semantic correspondence.
To address this, we introduce a very simple test-time pose alignment strategy that utilizes the global pose information inherent in deep features (\cref{subsec:global_pose_awareness}) and improves correspondence accuracy.

As in~\cref{fig:4.5pose}, we first augment the source image by using a set of pose-variant augmentations (\eg, flip, rotations \etc), calculate the IMD (\cref{eq:idm}) between the augmented source images and the target image, and choose the optimal pose with the minimum IMD distance.\footnote{Refer to Supp.~\ref{sec:alternative_pose_alignment} for alternative metrics that does not require mask.}
As in \cref{fig:4zeroshot_exp}, this simple pose alignment can drastically improve the correspondence accuracy in a test-time, unsupervised manner.

\subsection{Dense Training Objective}
\label{sec:dense_objective}

Let $\mathbf{F}$ represent the raw feature map and $\mathrm{f}(\cdot)$ be the post-processing model that outputs the refined feature map $\Tilde{\mathbf{F}} = \mathrm{f}(\mathbf{F})$, illustrated in \cref{fig:5supervised}.
Given a set of annotated keypoint pairs from source images $\mathcal{P}^s = \{\mathbf{p}_1^s, \mathbf{p}_2^s, \ldots, \mathbf{p}_n^s\}$ and target images $\mathcal{P}^t = \{\mathbf{p}_1^t, \mathbf{p}_2^t, \ldots, \mathbf{p}_n^t\}$, previous works with pretrained foundation model features~\cite{luo2023diffusion,zhang2023tale} adopt a CLIP-style symmetric contrastive loss $\mathrm{CL}(\cdot,\cdot)$ to train the post-processing model:
\begin{equation}
\mathcal{L}_\text{sparse} = \mathrm{CL}(\Tilde{\mathbf{F}}^s(\mathcal{P}^s), \Tilde{\mathbf{F}}^t(\mathcal{P}^t)),
\end{equation}
where $\Tilde{\mathbf{F}}^s$ and $\Tilde{\mathbf{F}}^t$ are the post-processed source and target features.
However, the loss is applied only to features with sparsely annotated keypoints, which potentially neglects additional informative features.

Instead, we employ the soft-argmax operator \cite{kendall2017end,lee2019sfnet,yang2019volumetric} so that gradients calculated from sparse annotations can be back-propagated to all spatial locations.
Specifically, we compute the similarity map $S_i = {\Tilde{\mathbf{F}}^s(\mathbf{p}_i^s)}^{T} \Tilde{\mathbf{F}}^t$ between the normalized query keypoint descriptor ${\Tilde{\mathbf{F}}^s(\mathbf{p}_i^s)}$ and the target feature map $\Tilde{\mathbf{F}}^t$.
Then, we take soft-argmax over the similarity map to get the predicted position $\hat{\mathbf{p}}_i^t = \operatorname{SoftArgmax}(S_i)$. 
The L2 norm penalizes the distance between the predicted position and the target position $\hat{\mathbf{p}}_i^t$:
\aroundeqn
\begin{equation}
\mathcal{L}_\text{dense} = \sum_{i} \| \hat{\mathbf{p}}_i^t - (\mathbf{p}_i^t + \epsilon) \|_2,
\end{equation}
\aroundeqn
To prevent overfitting, we also apply Dropout at the input feature map $\mathbf{F}$ and Gaussian noise $\epsilon$ that perturbs the ground truth keypoint positions $\mathbf{p}_i^t$.
We empirically find that combining the two objectives achieves better performance; thus our final training objective is $\mathcal{L} = \mathcal{L}_\text{dense} + \mathcal{L}_\text{sparse}$.

\begin{table*}[t]
\centering
\renewcommand{\arraystretch}{1.025}
\renewcommand{\tabcolsep}{1.5pt}
\caption{\textbf{Evaluation on SPair-71k.} Per-class and average PCK@$0.10$ on test split. 
The methods are categorized into two types: supervised (S) and unsupervised (U). 
${\dag}$: index is used to flip source keypoints at test time.
${*}$: fine-tuned backbone. 
We report \textit{per point} PCK result for the (U) methods, following~\cite{ofri2023neural, gupta2023asic}, and \textit{per image} result for the (S) methods, following~\cite{liu2020semantic, huang2022learning, lee2021patchmatch, cho2022cats++}.
The highest PCK are highlighted in \textbf{bold}, while the second highest are \underline{underlined}. Both our zero-shot and supervised methods outperform prior arts across all categories.
}

\aftertabcaption
\label{tab:spair_category}
\resizebox{1\textwidth}{!}{
\begin{tabular}{@{}clcccccccccccccccccccl@{}}
\toprule
 & {Method}  & Aero & Bike & Bird & Boat &Bottle&Bus&Car&Cat&Chair& Cow& Dog& Horse& Motor&Person& Plant& Sheep & Train & TV & \textbf{All}\\  
\midrule

\myrowcolour
\cellcolor{white}
\textbf{U}

& ASIC~\cite{gupta2023asic}                      & {57.9} & {25.2} & {68.1} &  {24.7}	& 35.4 & 28.4 & {30.9} &  {54.8}	& {21.6} & {45.0} &	{47.2} & {39.9} & {26.2} & {48.8} &	14.5 & {24.5} & {49.0} & 24.6 & {36.9}\\

& DINOv2+NN~\cite{oquab2023dinov2,zhang2023tale}             &{{72.7}} & {{62.0}} & {{85.2}} & {{41.3}} & {40.4} & {{52.3}} & {{51.5}} & {71.1} & {36.2} & {67.1} & {{64.6}} & {{67.6}} & {{61.0}} & {{68.2}} & {30.7} & {{62.0}} & {54.3} & {24.2} & {55.6} \\

\myrowcolour
\cellcolor{white}
& DIFT~\cite{tang2023dift}        & {63.5} & {54.5} & {80.8} & {34.5} & {46.2} & {52.7} & {48.3} & {77.7} & {39.0} & {76.0} & {54.9} & {61.3} & {53.3} & {46.0} & {57.8} & {57.1} & {71.1} & {63.4} & {57.7} \\

& SD+DINO~\cite{zhang2023tale}       & {{73.0}} & {{64.1}} & {{86.4}} & {{40.7}} & {{52.9}} & {{55.0}} & {{53.8}} & {{78.6}} & {{45.5}} & {{77.3}} & {{64.7}} & {{69.7}} & {{63.3}} & {{69.2}} & {{58.4}} & {{67.6}} & {{66.2}} & {{53.5}} & {{64.0}}\\
\cmidrule(){2-21} 

\myrowcolour
\cellcolor{white}
\textbf{U$^\dag$} & {NeuCongeal$^\dag$~\citep{ofri2023neural}} & {-} & {29.1} & {-} & {-} & {-} &  {-} & {-} & {53.3} & {-} & {-} & {35.2} & {-} & {-} & {-} & {-} & {-} & {-} & {-} & {-} \\

& \textbf{Ours-Zero-Shot}$^\dag$ & \textbf{78.0} & \textbf{66.4} & \textbf{90.2} & \textbf{44.5} & \textbf{60.1} & \textbf{66.6} & \textbf{60.8} & \textbf{82.7} & \textbf{53.2} & \textbf{82.3} & \textbf{69.5} & \textbf{75.1} & \textbf{66.1} & \textbf{71.7} & \textbf{58.9} & \textbf{71.6} & \textbf{83.8} & \textbf{55.5} & \textbf{69.6} \\

\hline
\hline

\myrowcolour
\cellcolor{white}
\textbf{S}
& SCOT~\citep{liu2020semantic}          & 34.9 & 20.7 & 63.8 & 21.1 & 43.5 & 27.3 & 21.3 & 63.1 & 20.0 & 42.9 & 42.5 & 31.1 & 29.8 & 35.0 & 27.7 & 24.4 & 48.4 & 40.8 & 35.6\\

& PMNC$^{*}$~\citep{lee2021patchmatch}        & 54.1 & 35.9 & 74.9 & 36.5 & 42.1 & 48.8 & 40.0 & 72.6 & 21.1 & 67.6 & 58.1 & 50.5 & 40.1 & 54.1 & 43.3 & 35.7 & 74.5 & 59.9 & 50.4\\

\myrowcolour
\cellcolor{white}
& SCorrSAN$^{*}$~\citep{huang2022learning}    & 57.1 & 40.3 & 78.3 & 38.1 & 51.8 & 57.8 & 47.1 & 67.9 & 25.2 & 71.3 & 63.9 & 49.3 & 45.3 & 49.8 & 48.8 & 40.3 & 77.7 & {69.7} & 55.3\\

& CATs++$^{*}$~\cite{cho2022cats++}    & 60.6 & 46.9 & 82.5 & 41.6 & {56.8} & 64.9 & 50.4 & 72.8 & 29.2 & 75.8 & 65.4 & 62.5 & 50.9 & 56.1 & 54.8 & 48.2 & 80.9 & {74.9} & 59.8\\

\myrowcolour
\cellcolor{white}
& DHF~\cite{luo2023diffusion}  & 74.0 & 61.0 & 87.2 & 40.7 & 47.8 & 70.0 & 74.4 & 80.9 & 38.5 & 76.1 & 60.9 & 66.8 & 66.6 & 70.3 & 58.0 & 54.3 & 87.4 & 60.3 & 64.9 \\

&   SD+DINO (S)~\cite{zhang2023tale}  & {81.2} & {66.9} & {91.6} & {61.4} & {57.4} & {85.3} & {83.1} & {90.8} & {54.5} & {88.5} & {75.1} & {80.2} & {71.9} & {77.9} & {60.7} & {68.9} & {92.4} & {65.8} & {74.6} \\
\cmidrule(){2-21} 

\myrowcolour
\cellcolor{white}
& \textbf{Ours}  & {87.0} & {73.7} &{95.4} & {69.0} & {66.1} & \underline{91.6} & {86.9} & {90.7} & \underline{68.6} & \underline{93.6} & {85.2} & {84.6} & {78.7} & {86.9} & {79.7} & \underline{79.0} & \textbf{96.9} & {84.3} & {82.9} \\

&\textbf{{Ours (Adapt. Pose)$^{\dag}$}} &
\underline{87.6} & \underline{74.1} & \underline{95.5} & \underline{70.1} & \underline{66.7} & \textbf{92.0} & \underline{87.4} & \underline{91.4} & 68.0 & 93.2 & \underline{85.5} & \underline{84.7} & \underline{79.9} & \underline{87.8} & \underline{79.9} & 78.9 & \textbf{96.9} & \underline{84.8} &
{\underline{83.2}} \\

\myrowcolour
\cellcolor{white}
& \textbf{Ours (AP-10K P.T.)}  & \textbf{92.0} & \textbf{76.1} & \textbf{97.2} & \textbf{70.4} & \textbf{70.5} & {91.4} & \textbf{89.7} & \textbf{92.7} & \textbf{73.4} & \textbf{95.0} & \textbf{90.5} & \textbf{87.7} & \textbf{81.8} & \textbf{91.6} & \textbf{82.3} & \textbf{83.4} & {96.5} & \textbf{85.3} &  \textbf{{85.6}} \\
\bottomrule
\end{tabular}
} %
\aftertab
\end{table*}

\vspace{-0.125em}
\subsection{Pose-variant Augmentation}
\label{sec:pa_augment}
\vspace{-0.125em}

Standard data augmentation schemes (\eg, random cropping, color jittering, \etc) have been generally used to augment the limited number of annotated data.
However, such standard augmentations show two shortcomings in naively adopting them to our approach.
Diverse augmentations on input images require our model to process the feature map of each augmented image using visual foundation models, which linearly increases the computational cost along with the number of augmentation schemes used.
Besides, such augmentations (\eg, cropping, scaling, or photometric augmentations) do not augment images with different poses or viewpoints, which might not bring additional effective supervision signals for geometric awareness.

Instead, we introduce a set of pose-varying augmentation schemes tailored to our approach, which needs to process only one feature map from a single additional augmented image (horizontal flipped) yet can utilize the feature in multiple ways. 
The motivation is that the deep features are aware of the global pose; thus, the processed feature map of the flipped image can add an additional signal; compared to simply flipping the feature map.
We introduce the following three augmentation settings: 1) \textit{double flip}: flipped source image and flipped target image; 2) \textit{single flip}: flipped source image and original target image; and 3) \textit{self flip}: source image and flipped source image.
For setting 2 and 3, keypoint annotations are correspondingly flipped to preserve the inherent geometric concept, \eg, the left paw in the flipped image should be the right paw of the original image.
The keypoint flipping in \textit{self flip} also ensures that the model learns to discern concepts rather than simply matching keypoints based on appearances.

\vspace{-0.125em}
\subsection{Window Soft Argmax}
\label{sec:window}
\vspace{-0.125em}

At test time, current methods~\cite{luo2023diffusion, zhang2023tale} use the argmax operation on the similarity map to infer correspondence.
However, it shows two major limitations: argmax is limited to discrete pixel coordinates without sub-pixel reasoning, and it does not incorporate any spatial context with neighboring pixels when determining correspondence.
One could use soft-argmax at time time too, but our study shows in~\cref{tab:ablation} that it does not improve the performance on all metrics, probably due to its nature of incorporating similarities from all pixels with possible noisy response.

To complement the weaknesses of both, we propose a \textit{window soft argmax} technique for both supervised and unsupervised settings.
First, we determine the target center location using the argmax operation and apply soft-argmax on the pre-defined window, as illustrated in \cref{fig:5supervised}.
This hybrid approach naturally enables sub-pixel reasoning but also prevents it from being affected by any noisy response in the similarity map.
\cref{tab:ablation} shows that the usage of window soft argmax substantially improves the correspondence performance on all metrics.

\begin{table*}[t]
\centering
\footnotesize
\renewcommand{\arraystretch}{1.0}
\caption{\textbf{Evaluation on SPair-71k, AP-10K, and PF-Pascal datasets at different PCK levels.} We report the performance of the AP-10K intra-species (I.S.), cross-species (C.S.), and cross-family (C.F.) test sets.
${\dag}$: index is used to flip source keypoints at test time. 
${*}$: fine-tuned backbone. 
We report the \textit{per image} PCK results (hence the (U) results are different from~\cref{tab:spair_category}). The highest and second PCK among each category is \textbf{bold} and \underline{underlined}, respectively. 
Both our zero-shot and supervised methods outperform all previous methods significantly.
}

\aftertabcaption
\label{tab:all_level_pck}
\resizebox{1\textwidth}{!}{
\begin{tabular}{@{}clccc|ccc|ccc|ccc|ccc@{}}
\toprule
&&\multicolumn{3}{c}{SPair-71k} &\multicolumn{3}{c}{AP-10K-I.S.} &\multicolumn{3}{c}{AP-10K-C.S.} &\multicolumn{3}{c}{AP-10K-C.F.} &\multicolumn{3}{c}{PF-Pascal}\\
\cmidrule(l){3-5} \cmidrule(l){6-8} \cmidrule(l){9-11} \cmidrule(l){12-14} \cmidrule(l){15-17} 
 & {Method}  & 0.01 & 0.05 & 0.10 & 0.01 & 0.05 & 0.10& 0.01 & 0.05 & 0.10& 0.01 & 0.05 & 0.10& 0.05 & 0.10 &0.15\\  
\midrule
\multirow{1}{0.125cm}{\textbf{U}}
& DINOv2+NN~\cite{oquab2023dinov2,zhang2023tale}       &6.3&38.4&53.9&6.4&41.0&60.9&5.3&37.0&57.3&4.4&29.4&47.4&63.0&79.2&85.1  \\
& DIFT~\cite{tang2023dift}       &7.2&39.7&52.9&6.2&34.8&50.3&5.1&30.8&46.0&3.7&22.4&35.0&66.0&81.1&87.2  \\
& SD+DINO~\cite{zhang2023tale}       &7.9&44.7&59.9&7.6&43.5&62.9&6.4&39.7&59.3&5.2&30.8&48.3&71.5&85.8&90.6  \\
\cmidrule(){2-17} 
\multirow{1}{0.125cm}{\textbf{U}$^\dag$}& {\textbf{Ours-Zero-Shot}$^\dag$} & \textbf{9.9} & \textbf{49.1} & \textbf{65.4} & \textbf{11.3} & \textbf{49.8} & \textbf{68.7} & \textbf{9.3} & \textbf{44.9} & \textbf{64.6} & \textbf{7.4} & \textbf{34.9} & \textbf{52.7} & \textbf{74.0} & \textbf{86.2} & \textbf{90.7} \\
\hline
\hline
\multirow{1}{0.125cm}{\textbf{S}}
&  SCorrSAN$^{*}$~\cite{huang2022learning}  &3.6&36.3&55.3&-&-&-&-&-&-&-&-&-&81.5&93.3&96.6  \\
&   CATs++$^{*}$~\cite{cho2022cats++}  &4.3&40.7&59.8&-&-&-&-&-&-&-&-&-&{84.9}&93.8&96.8  \\
&   DHF~\cite{luo2023diffusion}  &8.7&50.2&64.9&8.0&45.8&62.7&6.8&42.4&60.0&5.0&32.7&47.8&78.0&90.4&94.1  \\
&   SD+DINO (S)~\cite{zhang2023tale}  &9.6&57.7&74.6&{9.9}&{57.0}&{77.0}&{8.8}&{53.9}&{74.0}&{6.9}&{46.2}&{65.8}&80.9&93.6&96.9  \\
\cmidrule(){2-17} 
&  \textbf{Ours}  & {21.6}& {72.6}& {82.9}& \underline{23.1}& \underline{73.0}& \underline{87.5}&\textbf{21.7}&\underline{70.2}&\underline{85.8}&\textbf{18.4}&\underline{63.1}&\underline{78.4}&\underline{85.5}&\underline{95.1}&\underline{97.4} \\
& {\textbf{Ours (Adapt. Pose)$^{\dag}$}} & \underline{21.7} & \underline{72.8} & \underline{83.2} & \textbf{{23.2}} & \textbf{{73.2}} & \textbf{{87.7}} & \textbf{{21.7}} & \textbf{{70.3}} & \textbf{{85.9}} & \underline{{18.3}} & \textbf{{63.2}} & \textbf{{78.5}} & {85.3} & {95.0} & \underline{97.4} \\
&  \textbf{Ours (AP-10K P.T.)}  & \textbf{22.0}&\textbf{75.3} & \textbf{85.6}& -&-&-&-&-&-&-&-&-&\textbf{85.9}&\textbf{95.7}&\textbf{98.0}   \\
\bottomrule
\end{tabular}
}
\aftertab
\end{table*}

\begin{table}[t]
\centering
\renewcommand{\arraystretch}{1.0}
\caption{\textbf{Evaluation on the geometry-aware subset.} We report the results on both SPair-71k and AP-10K intra-species test sets across three PCK levels. The best performances are \textbf{bold}.}
\label{tab:geoaware_pck}

\aftertabcaption
\resizebox{0.475\textwidth}{!}{
\begin{tabular}{@{}clcccccc}
\toprule
&&\multicolumn{3}{c}{SPair-71k} &\multicolumn{3}{c}{AP-10K-I.S.}\\
\cmidrule(l){3-5} \cmidrule(l){6-8}
 & {Method} & 0.01 & 0.05 & 0.10 & 0.01 & 0.05 & 0.10\\
\midrule

\multirow{1}{0.125cm}{\textbf{U}}
&DINOv2+NN~\cite{oquab2023dinov2,zhang2023tale}&3.4&28.2&42.0&2.1&26.8&48.6\\
&DIFT~\cite{tang2023dift}&4.6&30.0&42.5&1.8&18.9&34.6\\
&SD+DINO~\cite{zhang2023tale}&5.3&34.5&49.3&2.5&28.0&49.5\\
\cmidrule(){2-8} 
\multirow{1}{0.125cm}{\textbf{U}$^\dag$}&{\textbf{Ours-Zero-Shot}$^\dag$}&\textbf{{6.9}}&\textbf{{39.5}}&\textbf{{56.8}}&\textbf{{3.5}}&\textbf{{35.9}}&\textbf{57.8}\\
\hline
\hline

\multirow{1}{0.125cm}{\textbf{S}}
& SCorrSAN$^{*}$~\cite{huang2022learning} &2.8&30.0&49.4&-&-&-\\
& CATs++$^{*}$~\cite{cho2022cats++} &3.2&33.1&53.0&-&-&-\\
& DHF~\cite{luo2023diffusion} &6.8&42.1&56.7&2.5&30.0&50.7\\
& SD+DINO (S)~\cite{zhang2023tale} &7.5&50.3&67.6&4.0&43.7&69.3\\
\cmidrule(){2-8} 
&\textbf{Ours}&18.2&66.0&77.4&{10.4}&{64.8}&{82.8}\\
&{\textbf{Ours (Adapt. Pose)$^{\dag}$}}&{{18.3}}&{{66.3}}&{{78.0}}&\textbf{{10.5}}&\textbf{{65.0}}&\textbf{83.2}\\
&\textbf{Ours (AP-10K P.T.)}&\textbf{20.1}&\textbf{71.0}&\textbf{82.3}&-&-&-\\
\bottomrule
\end{tabular}
}
\aftertab
\end{table}

\vspace{-0.125em}
\section{Experimental Results}
\label{sec:experiments}

\vspace{-0.25em}
\myparagraph{Implementation details.} 
We follow~\cite{zhang2023tale} to resize the input image to $960^2$ and $840^2$ to extract the SD and DINOv2 features, respectively, yielding a feature map at a resolution of $60\times 60$. 
The post-processor on top of the fused features is four bottleneck layers~\cite{he2016deep} with 5M parameters in total. 
The model is trained with the AdamW optimizer~\cite{loshchilov2017decoupled} of weight decay rate $0.001$ and the one-cycle scheduler~\cite{smith2019super} of $1.25 \times 10^{-3}$ learning rate and $0.3$ percentage for the increasing cycle.
We train all our models on one NVIDIA RTX3090 GPU. Refer to Supp.~\ref{sec:more_implementation_details} for more details.

\vspace{-0.125em}
\myparagraph{Datasets.} 
We evaluate our methods on two widely-used benchmarks, namely PF-Pascal and SPair-71k, and our new proposed benchmark.
\textit{PF-Pascal}~\cite{ham2016proposal} consists of 2941 training, 308 validation, and 299 testing image pairs with similar viewpoints and instance pose. The images span across 20 categories of objects.
\textit{SPair-71k}~\cite{min2019spair} is a more challenging and larger-scale dataset with $53,340$ training pairs, $5,384$ validation pairs, and $12,234$ testing pairs across 18 categories, with large intra-class variation.

\myparagraph{AP-10K benchmark.} 
To further validate and improve our method in an in-the-wild setting, we build a new large-scale, challenging semantic correspondence benchmark with an existing animal pose estimation dataset.
The AP-10K dataset~\cite{yu2021ap} consists of 10,015 images across 23 families and 54 species. 
All the images share the same keypoint annotation of 17 keypoints.
After manually filtering images with multiple instances and images with less than three visible keypoints, we construct a benchmark with 261k training, 17k validation, and 36k testing image pairs.
The validation and testing image pairs span three settings: the main intra-species set, the cross-species set, and the cross-family set.
It is 5× larger than the largest existing benchmark~\cite{min2019spair}.
Please refer to Supp.~\ref{sec:benchmark} for more details.

\myparagraph{Evaluation metrics.} We follow the common practice and use the Percentage of Correct Keypoints
(PCK)~\cite{yang2012articulated} to evaluate the correspondence accuracy. 
The PCK is computed within a threshold of $\alpha \cdot max(h, w)$ where $\alpha$ is a positive decimal (\eg, 0.10) and $(h, w)$ denotes the dimensions of the bounding box of an instance in SPair-71k and AP-10K, and the dimensions of the images in PF-Pascal, respectively.

\subsection{Quantitative Analysis}

\vspace{-0.25em}
\myparagraph{Overall semantic correspondence.}
The per-category evaluation results, presented in~\cref{tab:spair_category}, demonstrate the efficacy of our methods. 
Our \textit{zero-shot} approach, despite its simplicity, achieves considerable gains over SD+DINO, highlighting the significance of pose alignment in semantic correspondence. 
In the \textit{supervised} category, our methods outperform existing works across all 18 categories, registering a substantial improvement of \textbf{11.0p} (from 74.6 to 85.6). 
Notably, pre-training on the AP-10K dataset contributes a gain of 2.7p, underscoring the untapped potential of animal pose datasets in this domain.

Further comparisons across different datasets and three PCK levels are in~\cref{tab:all_level_pck}. 
Our methods exhibit significant improvements across most metrics, particularly with notable gains in the more strict thresholds (e.g., PCK@0.05 and PCK@0.01), especially considering that SD+DINO uses the same raw feature maps as our model. 
Despite the methods being trained only on AP-10K intra-species sets, the robust performance on cross-species and cross-family test sets showcases the generalizability of our approach.

\begin{figure*}[t]
  \centering
   \includegraphics[width=0.9\linewidth]{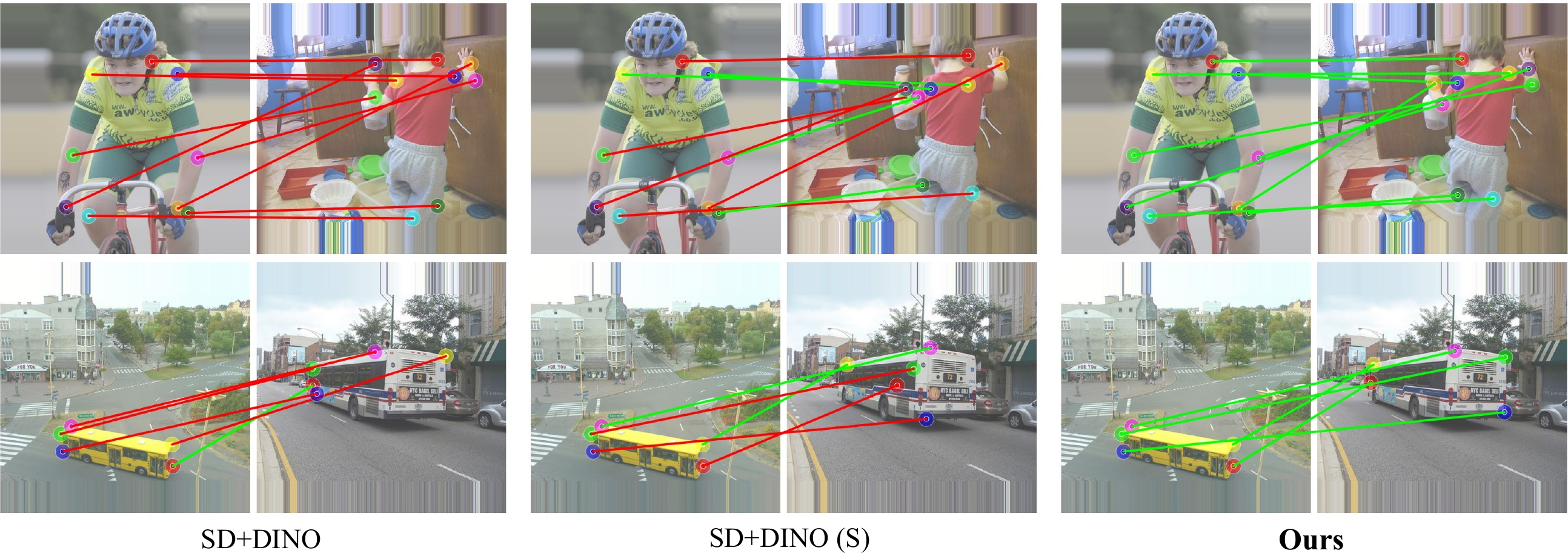}
   \vspace{-0.85em}
   \caption{\textbf{Qualitative comparison.} 
   \green{Green lines} indicate correct matches and \red{red} incorrect.
   Our method can build geometrically correct semantic correspondence even at extreme view variation, while both versions of SD+DINO  struggle with geometric ambiguity (\eg, ear and hands in the person example, corners in the bus example).
   Please refer to Supp.~\ref{sec:qualitative_ap10k} and~\ref{sec:qualitative_spair} for more results.
   }
   \label{fig:7sparse}
   \afterfig
\end{figure*}

\begin{figure*}[t]
  \centering
   \includegraphics[width=1\linewidth]{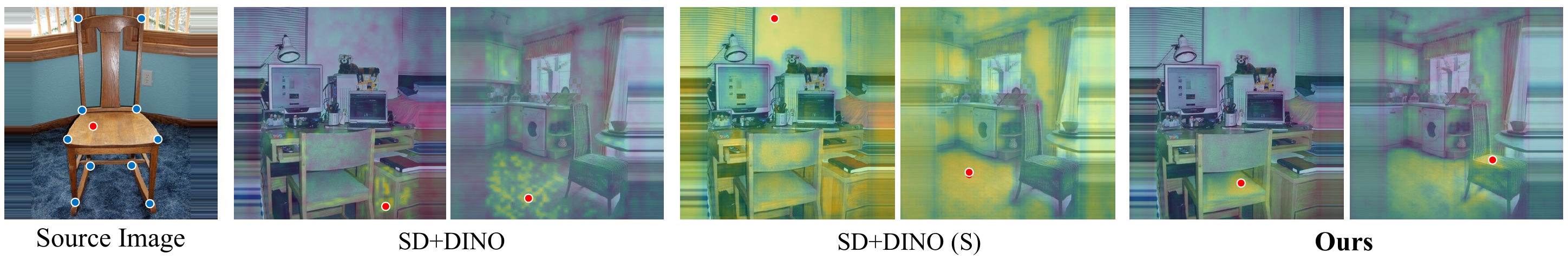}
   \vspace{-2em}
   \caption{\textbf{Visualization of the similarity map.} For the \red{red} query point, SD+DINO matches appearance-similar points (wooden desk, floor); SD+DINO (S) returns a noisy similarity map due to the query point being out of supervision. Our method locates both semantically and geometrically correct points. The keypoint supervision of ``chair" category is in \blue{blue}, though these images are not in the training set. %
   }
   \label{fig:7similarity}
   \afterfig
   \vspace{-0.5em}
\end{figure*}

\myparagraph{Geometry-aware semantic correspondence.}
Our methods achieve even more significant improvements in the geometry-aware subset, as reported in~\cref{tab:geoaware_pck}. 
We reduce the relative gap from $9.38\%$ (SD+DINO (S)) to $3.86\%$ on the SPair-71k PCK@0.10 metric.
Notably, the proposed adaptive viewpoint alignment brings more substantial gain on the geometry-aware subset for both zero-shot and supervised settings, suggesting its effectiveness in improving the geometric ambiguity by mitigating the pose variation.
Besides, pre-training on the AP-10K dataset brings even a gain of 4.3p on the geo-aware subset.

\myparagraph{Ablation studies.}
\begin{table}[t]
\centering
\renewcommand{\arraystretch}{1.015}
\caption{\textbf{Ablation study on SPair-71k.} We report the PCK@$\alpha_\text{bbox}$ results for both standard set (Std.) and geometry-aware set (Geo.). The best performances are \textbf{bold}. Our default method is \underline{underlined}.}
\label{tab:ablation}
\aftertabcaption
\resizebox{0.475\textwidth}{!}{
\begin{tabular}{rlcccccc}
\toprule
&&\multicolumn{3}{c}{SPair-71k (Std.)} &\multicolumn{3}{c}{SPair-71k (Geo.)}\\
\cmidrule(l){3-5} \cmidrule(l){6-8}
  &{Model Variants} & 0.01 & 0.05 & 0.10 & 0.01 & 0.05 & 0.10\\
\midrule

&Baseline&9.6& 57.7& 74.6& 7.5& 50.3& 67.6\\
\textbf{+}&Dense Training Objective&13.0& 65.2& 78.3& 11.1& 58.8& 71.9\\
\textbf{+}&Pose-variant Augmentation&13.8& 66.7& 80.0& 11.4& 60.5& 73.9\\
\textbf{+}&Perturbation \& Dropout&15.1& 69.3& 81.3& 13.5& 63.3& 75.4\\
\hline
\multirow{4}{*}{\textbf{+}}
&Soft Argmax Inference &20.5& 69.6& 81.0& 16.9& 61.9& 75.0\\
&Window Soft Argmax (5)&\textbf{22.3}& 72.1& 82.0& \textbf{19.8}& 66.0& 76.5\\
&Window Soft Argmax (9)&22.0& \textbf{72.7}& 82.5& 19.2& \textbf{66.3}& 77.1\\
&\underline{Window Soft Argmax (15)}&21.6& 72.6& \textbf{82.9}& 18.2& 66.0& \textbf{77.4}\\
\bottomrule
\end{tabular}
}
\aftertab
\end{table}
Further ablation studies are in~\cref{tab:ablation}. 
Each element of our designs brings about moderate improvements. 
The dense training objective, pose-variant augmentation, and window soft argmax notably enhance results in the geometry-aware subset, while ground truth perturbation and feature map Dropout improve the overall correspondence (as shown in the similar gain on both sets).
Regarding the window soft argmax, varying window sizes have different effects across three thresholds. 
We set the window size to $15 \times 15$ and $11 \times 11$ for the supervised and unsupervised setting respectively, for the optimal balance.

In the Supp.~\ref{sec:leave-one-out}, we also provide a leave-one-out ablation study with the breakdown evaluation protocol introduced in~\cite{aygun2022demystifying}, to evaluate the detailed effect of each of our proposed module. 
In short, all our designs expect pertubation\&dropout notably improves the geometry-aware (\eg, left/right) confusion, while the dense training objective also reduces mismatches to the image background.

\subsection{Qualitative Analysis}

We qualitatively compare our methods against both zero-shot and supervised versions of SD+DINO. 
As shown in~\cref{fig:7sparse}, our approach significantly enhances semantic correspondence under the extreme view-variation cases.
While additional supervision in SD+DINO does aid in keypoint localization to some extent, both versions of SD+DINO struggle with geometric ambiguity.

We further investigate cases where the query point lacks meaning and without direct supervision.
As the visualization of the similarity map shown in the~\cref{fig:7similarity}, SD+DINO highlights the regions with similar appearance (wooden materials) but fails to locate the chair; SD+DINO (S) generates noisy similarity maps (all regions are highlighted) when the query point is out of supervision, due to the sparse training objective; Our method locates the points both semantically and geometrically correct.
Despite all methods sharing the same raw feature maps and our approach using the same feature post-processor as SD+DINO (S), the improvements in our method underscore the effectiveness of our design.

\begin{figure}[t]
  \centering
   \includegraphics[width=1\linewidth]{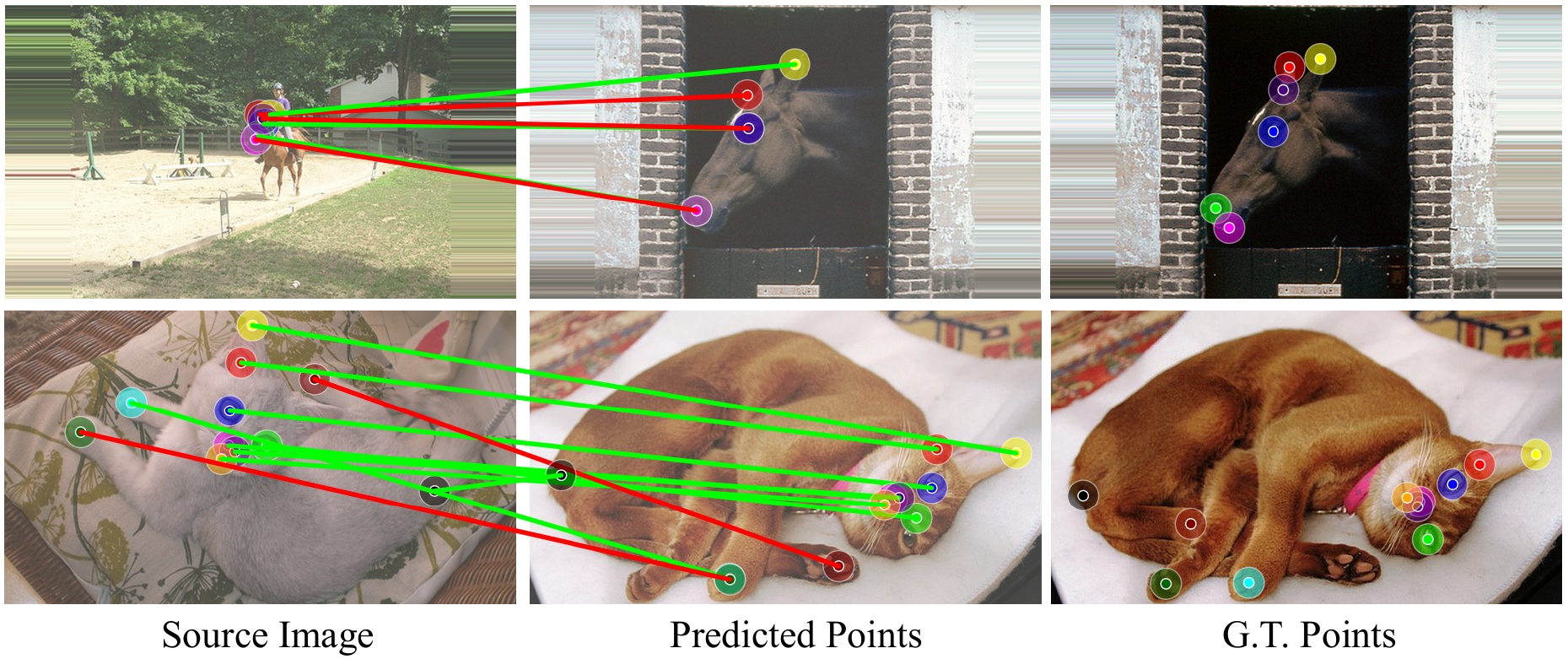}
   \vspace{-2em}
   \caption{\textbf{Limitations.} Top: small instance. Bottom: scenarios combining both large pose variation and severe deformation.}
   \label{fig:7limitation}
   \vspace{-1.5em}
\end{figure}

\myparagraph{Limitations.}
As shown in~\cref{fig:7limitation} (top), small instances may be challenging for our method due to the resolution limits of raw feature maps. %
Our method may fail on extreme pose variations with severe deformation (see~\cref{fig:7limitation}, bottom). Future work may address these complex scenarios by advanced reasoning mechanisms or geometric prior, such as the spherical constraint proposed in concurrent work~\cite{mariotti2023improving}.

\section{Conclusion}
\label{sec:conclusion}
We identified the problem of geometry ambiguity in semantic correspondence and introduced simple and effective techniques to improve current methods.
We also developed a new benchmark to train and validate existing methods.
Extensive experiments demonstrate that our method not only significantly improves the overall semantic correspondence but also narrows the gap between the geometry-aware sub-set and the standard set, thereby benefiting various downstream tasks and providing another angle to understand the internal representation of foundation models.

\clearpage
{
    \small
    \bibliographystyle{ieeenat_fullname}
    \bibliography{main}

\begin{thebibliography}{62}
\providecommand{\natexlab}[1]{#1}
\providecommand{\url}[1]{\texttt{#1}}
\expandafter\ifx\csname urlstyle\endcsname\relax
  \providecommand{\doi}[1]{doi: #1}\else
  \providecommand{\doi}{doi: \begingroup \urlstyle{rm}\Url}\fi

\bibitem[Aberman et~al.(2018)Aberman, Liao, Shi, Lischinski, Chen, and Cohen-Or]{aberman2018neural}
Kfir Aberman, Jing Liao, Mingyi Shi, Dani Lischinski, Baoquan Chen, and Daniel Cohen-Or.
\newblock Neural best-buddies: Sparse cross-domain correspondence.
\newblock \emph{ACM TOG}, 37\penalty0 (4):\penalty0 1--14, 2018.

\bibitem[Amir et~al.(2021)Amir, Gandelsman, Bagon, and Dekel]{amir2021deep}
Shir Amir, Yossi Gandelsman, Shai Bagon, and Tali Dekel.
\newblock Deep vit features as dense visual descriptors.
\newblock \emph{arXiv preprint arXiv:2112.05814}, 2\penalty0 (3):\penalty0 4, 2021.

\bibitem[Ayg{\"u}n and Mac~Aodha(2022)]{aygun2022demystifying}
Mehmet Ayg{\"u}n and Oisin Mac~Aodha.
\newblock Demystifying unsupervised semantic correspondence estimation.
\newblock In \emph{ECCV}, pages 125--142. Springer, 2022.

\bibitem[Banik et~al.(2021)Banik, Li, and Dong]{banik2021novel}
Prianka Banik, Lin Li, and Xishuang Dong.
\newblock A novel dataset for keypoint detection of quadruped animals from images.
\newblock \emph{arXiv preprint arXiv:2108.13958}, 2021.

\bibitem[Caron et~al.(2021)Caron, Touvron, Misra, J{\'e}gou, Mairal, Bojanowski, and Joulin]{caron2021emerging}
Mathilde Caron, Hugo Touvron, Ishan Misra, Herv{\'e} J{\'e}gou, Julien Mairal, Piotr Bojanowski, and Armand Joulin.
\newblock Emerging properties in self-supervised vision transformers.
\newblock In \emph{ICCV}, pages 9650--9660, 2021.

\bibitem[Cho et~al.(2021)Cho, Hong, Jeon, Lee, Sohn, and Kim]{cho2021cats}
Seokju Cho, Sunghwan Hong, Sangryul Jeon, Yunsung Lee, Kwanghoon Sohn, and Seungryong Kim.
\newblock Cats: Cost aggregation transformers for visual correspondence.
\newblock In \emph{NeurIPS}, pages 9011--9023, 2021.

\bibitem[Cho et~al.(2022)Cho, Hong, and Kim]{cho2022cats++}
Seokju Cho, Sunghwan Hong, and Seungryong Kim.
\newblock Cats++: Boosting cost aggregation with convolutions and transformers.
\newblock \emph{IEEE TPAMI}, 2022.

\bibitem[Dalal and Triggs(2005)]{dalal2005histograms}
Navneet Dalal and Bill Triggs.
\newblock Histograms of oriented gradients for human detection.
\newblock In \emph{CVPR}, pages 886--893. Ieee, 2005.

\bibitem[Geyer et~al.(2023)Geyer, Bar-Tal, Bagon, and Dekel]{geyer2023tokenflow}
Michal Geyer, Omer Bar-Tal, Shai Bagon, and Tali Dekel.
\newblock Tokenflow: Consistent diffusion features for consistent video editing.
\newblock \emph{arXiv preprint arXiv:2307.10373}, 2023.

\bibitem[Gupta et~al.(2023)Gupta, Jampani, Esteves, Shrivastava, Makadia, Snavely, and Kar]{gupta2023asic}
Kamal Gupta, Varun Jampani, Carlos Esteves, Abhinav Shrivastava, Ameesh Makadia, Noah Snavely, and Abhishek Kar.
\newblock Asic: Aligning sparse in-the-wild image collections.
\newblock In \emph{ICCV}, 2023.

\bibitem[Ham et~al.(2016)Ham, Cho, Schmid, and Ponce]{ham2016proposal}
Bumsub Ham, Minsu Cho, Cordelia Schmid, and Jean Ponce.
\newblock Proposal flow.
\newblock In \emph{CVPR}, pages 3475--3484, 2016.

\bibitem[Ham et~al.(2017)Ham, Cho, Schmid, and Ponce]{ham2017proposal}
Bumsub Ham, Minsu Cho, Cordelia Schmid, and Jean Ponce.
\newblock Proposal flow: Semantic correspondences from object proposals.
\newblock \emph{IEEE TPAMI}, 40\penalty0 (7):\penalty0 1711--1725, 2017.

\bibitem[He et~al.(2016)He, Zhang, Ren, and Sun]{he2016deep}
Kaiming He, Xiangyu Zhang, Shaoqing Ren, and Jian Sun.
\newblock Deep residual learning for image recognition.
\newblock In \emph{CVPR}, pages 770--778, 2016.

\bibitem[Hedlin et~al.(2023)Hedlin, Sharma, Mahajan, Isack, Kar, Tagliasacchi, and Yi]{hedlin2023unsupervised}
Eric Hedlin, Gopal Sharma, Shweta Mahajan, Hossam Isack, Abhishek Kar, Andrea Tagliasacchi, and Kwang~Moo Yi.
\newblock Unsupervised semantic correspondence using stable diffusion.
\newblock In \emph{NeurIPS}, 2023.

\bibitem[Hong et~al.(2022)Hong, Cho, Nam, Lin, and Kim]{hong2022cost}
Sunghwan Hong, Seokju Cho, Jisu Nam, Stephen Lin, and Seungryong Kim.
\newblock Cost aggregation with 4{D} convolutional swin transformer for few-shot segmentation.
\newblock In \emph{ECCV}, pages 108--126. Springer, 2022.

\bibitem[Huang et~al.(2022)Huang, Yang, He, Zhang, He, and Shrivastava]{huang2022learning}
Shuaiyi Huang, Luyu Yang, Bo He, Songyang Zhang, Xuming He, and Abhinav Shrivastava.
\newblock Learning semantic correspondence with sparse annotations.
\newblock In \emph{ECCV}, pages 267--284. Springer, 2022.

\bibitem[Huang et~al.(2023)Huang, Sun, Lai, Xu, Wang, Shen, and Ge]{huang2023weakly}
Yiwen Huang, Yixuan Sun, Chenghang Lai, Qing Xu, Xiaomei Wang, Xuli Shen, and Weifeng Ge.
\newblock Weakly supervised learning of semantic correspondence through cascaded online correspondence refinement.
\newblock In \emph{ICCV}, pages 16254--16263, 2023.

\bibitem[Hung et~al.(2019)Hung, Jampani, Liu, Molchanov, Yang, and Kautz]{hung2019scops}
Wei-Chih Hung, Varun Jampani, Sifei Liu, Pavlo Molchanov, Ming-Hsuan Yang, and Jan Kautz.
\newblock Scops: Self-supervised co-part segmentation.
\newblock In \emph{CVPR}, pages 869--878, 2019.

\bibitem[Kendall et~al.(2017)Kendall, Martirosyan, Dasgupta, Henry, Kennedy, Bachrach, and Bry]{kendall2017end}
Alex Kendall, Hayk Martirosyan, Saumitro Dasgupta, Peter Henry, Ryan Kennedy, Abraham Bachrach, and Adam Bry.
\newblock End-to-end learning of geometry and context for deep stereo regression.
\newblock In \emph{ICCV}, pages 66--75, 2017.

\bibitem[Kim et~al.(2018)Kim, Lin, Jeon, Min, and Sohn]{kim2018recurrent}
Seungryong Kim, Stephen Lin, Sang~Ryul Jeon, Dongbo Min, and Kwanghoon Sohn.
\newblock Recurrent transformer networks for semantic correspondence.
\newblock In \emph{NeurIPS}, 2018.

\bibitem[Kim et~al.(2019)Kim, Min, Jeong, Kim, Jeon, and Sohn]{kim2019semantic}
Seungryong Kim, Dongbo Min, Somi Jeong, Sunok Kim, Sangryul Jeon, and Kwanghoon Sohn.
\newblock Semantic attribute matching networks.
\newblock In \emph{CVPR}, pages 12339--12348, 2019.

\bibitem[Kirillov et~al.(2023)Kirillov, Mintun, Ravi, Mao, Rolland, Gustafson, Xiao, Whitehead, Berg, Lo, et~al.]{kirillov2023segment}
Alexander Kirillov, Eric Mintun, Nikhila Ravi, Hanzi Mao, Chloe Rolland, Laura Gustafson, Tete Xiao, Spencer Whitehead, Alexander~C Berg, Wan-Yen Lo, et~al.
\newblock Segment anything.
\newblock In \emph{ICCV}, pages 4015--4026, 2023.

\bibitem[Lee et~al.(2019)Lee, Kim, Ponce, and Ham]{lee2019sfnet}
Junghyup Lee, Dohyung Kim, Jean Ponce, and Bumsub Ham.
\newblock S{FN}et: Learning object-aware semantic correspondence.
\newblock In \emph{CVPR}, pages 2278--2287, 2019.

\bibitem[Lee et~al.(2020)Lee, Kim, Lee, Kim, Chang, and Choo]{lee2020reference}
Junsoo Lee, Eungyeup Kim, Yunsung Lee, Dongjun Kim, Jaehyuk Chang, and Jaegul Choo.
\newblock Reference-based sketch image colorization using augmented-self reference and dense semantic correspondence.
\newblock In \emph{CVPR}, pages 5801--5810, 2020.

\bibitem[Lee et~al.(2021)Lee, DeGol, Fragoso, and Sinha]{lee2021patchmatch}
Jae~Yong Lee, Joseph DeGol, Victor Fragoso, and Sudipta~N. Sinha.
\newblock Patchmatch-based neighborhood consensus for semantic correspondence.
\newblock In \emph{CVPR}, pages 13153--13163, 2021.

\bibitem[Liu et~al.(2020)Liu, Zhu, Yamada, and Yang]{liu2020semantic}
Yanbin Liu, Linchao Zhu, Makoto Yamada, and Yi Yang.
\newblock Semantic correspondence as an optimal transport problem.
\newblock In \emph{CVPR}, pages 4463--4472, 2020.

\bibitem[Long et~al.(2014)Long, Zhang, and Darrell]{long2014convnets}
Jonathan~L. Long, Ning Zhang, and Trevor Darrell.
\newblock Do convnets learn correspondence?
\newblock In \emph{NeurIPS}, 2014.

\bibitem[Loshchilov and Hutter(2019)]{loshchilov2017decoupled}
Ilya Loshchilov and Frank Hutter.
\newblock Decoupled weight decay regularization.
\newblock In \emph{ICLR}, 2019.

\bibitem[Lowe(1999)]{lowe1999object}
David~G. Lowe.
\newblock Object recognition from local scale-invariant features.
\newblock In \emph{ICCV}, pages 1150--1157. Ieee, 1999.

\bibitem[Luo et~al.(2023)Luo, Dunlap, Park, Holynski, and Darrell]{luo2023diffusion}
Grace Luo, Lisa Dunlap, Dong~Huk Park, Aleksander Holynski, and Trevor Darrell.
\newblock Diffusion hyperfeatures: Searching through time and space for semantic correspondence.
\newblock In \emph{NeurIPS}, 2023.

\bibitem[Mariotti et~al.(2023)Mariotti, Mac~Aodha, and Bilen]{mariotti2023improving}
Octave Mariotti, Oisin Mac~Aodha, and Hakan Bilen.
\newblock Improving semantic correspondence with viewpoint-guided spherical maps.
\newblock \emph{arXiv preprint arXiv:2312.13216}, 2023.

\bibitem[Min et~al.(2019)Min, Lee, Ponce, and Cho]{min2019spair}
Juhong Min, Jongmin Lee, Jean Ponce, and Minsu Cho.
\newblock Spair-71k: A large-scale benchmark for semantic correspondence.
\newblock \emph{arXiv prepreint arXiv:1908.10543}, 2019.

\bibitem[Min et~al.(2020)Min, Lee, Ponce, and Cho]{min2020learning}
Juhong Min, Jongmin Lee, Jean Ponce, and Minsu Cho.
\newblock Learning to compose hypercolumns for visual correspondence.
\newblock In \emph{ECCV}, pages 346--363. Springer, 2020.

\bibitem[Mou et~al.(2023)Mou, Wang, Song, Shan, and Zhang]{mou2023dragondiffusion}
Chong Mou, Xintao Wang, Jiechong Song, Ying Shan, and Jian Zhang.
\newblock Dragon{D}iffusion: Enabling drag-style manipulation on diffusion models.
\newblock \emph{arXiv preprint arXiv:2307.02421}, 2023.

\bibitem[Ofri-Amar et~al.(2023)Ofri-Amar, Geyer, Kasten, and Dekel]{ofri2023neural}
Dolev Ofri-Amar, Michal Geyer, Yoni Kasten, and Tali Dekel.
\newblock Neural congealing: Aligning images to a joint semantic atlas.
\newblock In \emph{CVPR}, pages 19403--19412, 2023.

\bibitem[Oquab et~al.(2023)Oquab, Darcet, Moutakanni, Vo, Szafraniec, Khalidov, Fernandez, Haziza, Massa, El-Nouby, et~al.]{oquab2023dinov2}
Maxime Oquab, Timoth{\'e}e Darcet, Th{\'e}o Moutakanni, Huy Vo, Marc Szafraniec, Vasil Khalidov, Pierre Fernandez, Daniel Haziza, Francisco Massa, Alaaeldin El-Nouby, et~al.
\newblock Dinov2: Learning robust visual features without supervision.
\newblock \emph{arXiv preprint arXiv:2304.07193}, 2023.

\bibitem[Peebles et~al.(2022)Peebles, Zhu, Zhang, Torralba, Efros, and Shechtman]{peebles2022gan}
William Peebles, Jun-Yan Zhu, Richard Zhang, Antonio Torralba, Alexei~A. Efros, and Eli Shechtman.
\newblock Gan-supervised dense visual alignment.
\newblock In \emph{CVPR}, pages 13470--13481, 2022.

\bibitem[Rocco et~al.(2017)Rocco, Arandjelovi{\'c}, and Sivic]{rocco2017convolutional}
Ignacio Rocco, Relja Arandjelovi{\'c}, and Josef Sivic.
\newblock Convolutional neural network architecture for geometric matching.
\newblock In \emph{CVPR}, pages 6148--6157, 2017.

\bibitem[Rocco et~al.(2018)Rocco, Arandjelovi{\'c}, and Sivic]{rocco2018end}
Ignacio Rocco, Relja Arandjelovi{\'c}, and Josef Sivic.
\newblock End-to-end weakly-supervised semantic alignment.
\newblock In \emph{CVPR}, pages 6917--6925, 2018.

\bibitem[Rombach et~al.(2022)Rombach, Blattmann, Lorenz, Esser, and Ommer]{rombach2022high}
Robin Rombach, Andreas Blattmann, Dominik Lorenz, Patrick Esser, and Bj{\"o}rn Ommer.
\newblock High-resolution image synthesis with latent diffusion models.
\newblock In \emph{CVPR}, pages 10684--10695, 2022.

\bibitem[Saharia et~al.(2022)Saharia, Chan, Saxena, Li, Whang, Denton, Ghasemipour, Gontijo~Lopes, Karagol~Ayan, Salimans, et~al.]{saharia2022photorealistic}
Chitwan Saharia, William Chan, Saurabh Saxena, Lala Li, Jay Whang, Emily~L. Denton, Kamyar Ghasemipour, Raphael Gontijo~Lopes, Burcu Karagol~Ayan, Tim Salimans, et~al.
\newblock Photorealistic text-to-image diffusion models with deep language understanding.
\newblock In \emph{NeurIPS}, pages 36479--36494, 2022.

\bibitem[Seo et~al.(2018)Seo, Lee, Jung, Han, and Cho]{seo2018attentive}
Paul~Hongsuck Seo, Jongmin Lee, Deunsol Jung, Bohyung Han, and Minsu Cho.
\newblock Attentive semantic alignment with offset-aware correlation kernels.
\newblock In \emph{ECCV}, pages 349--364, 2018.

\bibitem[Shtedritski et~al.(2023)Shtedritski, Vedaldi, and Rupprecht]{shtedritski2023learning}
Aleksandar Shtedritski, Andrea Vedaldi, and Christian Rupprecht.
\newblock Learning universal semantic correspondences with no supervision and automatic data curation.
\newblock In \emph{ICCV Workshops}, pages 933--943, 2023.

\bibitem[Simonyan and Zisserman(2015)]{simonyan2014very}
Karen Simonyan and Andrew Zisserman.
\newblock Very deep convolutional networks for large-scale image recognition.
\newblock In \emph{ICLR}, 2015.

\bibitem[Smith and Topin(2019)]{smith2019super}
Leslie~N. Smith and Nicholay Topin.
\newblock Super-convergence: Very fast training of neural networks using large learning rates.
\newblock In \emph{Artificial intelligence and machine learning for multi-domain operations applications}, pages 369--386. SPIE, 2019.

\bibitem[Tang et~al.(2023)Tang, Jia, Wang, Phoo, and Hariharan]{tang2023dift}
Luming Tang, Menglin Jia, Qianqian Wang, Cheng~Perng Phoo, and Bharath Hariharan.
\newblock Emergent correspondence from image diffusion.
\newblock In \emph{NeurIPS}, 2023.

\bibitem[Taniai et~al.(2016)Taniai, Sinha, and Sato]{taniai2016joint}
Tatsunori Taniai, Sudipta~N. Sinha, and Yoichi Sato.
\newblock Joint recovery of dense correspondence and cosegmentation in two images.
\newblock In \emph{CVPR}, pages 4246--4255, 2016.

\bibitem[Thewlis et~al.(2017)Thewlis, Bilen, and Vedaldi]{thewlis2017unsupervised}
James Thewlis, Hakan Bilen, and Andrea Vedaldi.
\newblock Unsupervised learning of object frames by dense equivariant image labelling.
\newblock 30, 2017.

\bibitem[Tian et~al.(2023)Tian, Aggarwal, Colaco, Kira, and Gonzalez-Franco]{tian2023diffuse}
Junjiao Tian, Lavisha Aggarwal, Andrea Colaco, Zsolt Kira, and Mar Gonzalez-Franco.
\newblock Diffuse, attend, and segment: Unsupervised zero-shot segmentation using stable diffusion.
\newblock \emph{arXiv preprint arXiv:2308.12469}, 2023.

\bibitem[Truong et~al.(2020{\natexlab{a}})Truong, Danelljan, Gool, and Timofte]{truong2020gocor}
Prune Truong, Martin Danelljan, Luc~V. Gool, and Radu Timofte.
\newblock Gocor: Bringing globally optimized correspondence volumes into your neural network.
\newblock In \emph{NeurIPS}, pages 14278--14290, 2020{\natexlab{a}}.

\bibitem[Truong et~al.(2020{\natexlab{b}})Truong, Danelljan, and Timofte]{truong2020glu}
Prune Truong, Martin Danelljan, and Radu Timofte.
\newblock Glu-net: Global-local universal network for dense flow and correspondences.
\newblock In \emph{CVPR}, pages 6258--6268, 2020{\natexlab{b}}.

\bibitem[Truong et~al.(2021{\natexlab{a}})Truong, Danelljan, Van~Gool, and Timofte]{truong2021learning}
Prune Truong, Martin Danelljan, Luc Van~Gool, and Radu Timofte.
\newblock Learning accurate dense correspondences and when to trust them.
\newblock In \emph{CVPR}, pages 5714--5724, 2021{\natexlab{a}}.

\bibitem[Truong et~al.(2021{\natexlab{b}})Truong, Danelljan, Yu, and Van~Gool]{truong2021warp}
Prune Truong, Martin Danelljan, Fisher Yu, and Luc Van~Gool.
\newblock Warp consistency for unsupervised learning of dense correspondences.
\newblock In \emph{ICCV}, pages 10346--10356, 2021{\natexlab{b}}.

\bibitem[Truong et~al.(2022)Truong, Danelljan, Yu, and Van~Gool]{truong2022probabilistic}
Prune Truong, Martin Danelljan, Fisher Yu, and Luc Van~Gool.
\newblock Probabilistic warp consistency for weakly-supervised semantic correspondences.
\newblock In \emph{CVPR}, pages 8708--8718, 2022.

\bibitem[Wah et~al.(2011)Wah, Branson, Welinder, Perona, and Belongie]{wah2011caltech}
Catherine Wah, Steve Branson, Peter Welinder, Pietro Perona, and Serge Belongie.
\newblock The {C}altech-{UCSD} birds-200-2011 dataset.
\newblock 2011.

\bibitem[Xu et~al.(2023)Xu, Liu, Vahdat, Byeon, Wang, and De~Mello]{xu2023open}
Jiarui Xu, Sifei Liu, Arash Vahdat, Wonmin Byeon, Xiaolong Wang, and Shalini De~Mello.
\newblock Open-vocabulary panoptic segmentation with text-to-image diffusion models.
\newblock In \emph{CVPR}, pages 2955--2966, 2023.

\bibitem[Yang et~al.(2017)Yang, Li, Cheng, Li, and Chen]{yang2017object}
Fan Yang, Xin Li, Hong Cheng, Jianping Li, and Leiting Chen.
\newblock Object-aware dense semantic correspondence.
\newblock In \emph{CVPR}, pages 2777--2785, 2017.

\bibitem[Yang and Ramanan(2019)]{yang2019volumetric}
Gengshan Yang and Deva Ramanan.
\newblock Volumetric correspondence networks for optical flow.
\newblock \emph{NeurIPS}, 32, 2019.

\bibitem[Yang and Ramanan(2012)]{yang2012articulated}
Yi Yang and Deva Ramanan.
\newblock Articulated human detection with flexible mixtures of parts.
\newblock \emph{IEEE TPAMI}, 35\penalty0 (12):\penalty0 2878--2890, 2012.

\bibitem[Yu et~al.(2021)Yu, Xu, Zhang, Zhao, Guan, and Tao]{yu2021ap}
Hang Yu, Yufei Xu, Jing Zhang, Wei Zhao, Ziyu Guan, and Dacheng Tao.
\newblock Ap-10k: A benchmark for animal pose estimation in the wild.
\newblock In \emph{NeurIPS}, 2021.

\bibitem[Zhang et~al.(2023)Zhang, Herrmann, Hur, Cabrera, Jampani, Sun, and Yang]{zhang2023tale}
Junyi Zhang, Charles Herrmann, Junhwa Hur, Luisa~Polania Cabrera, Varun Jampani, Deqing Sun, and Ming-Hsuan Yang.
\newblock A tale of two features: Stable diffusion complements dino for zero-shot semantic correspondence.
\newblock In \emph{NeurIPS}, 2023.

\bibitem[Zhao et~al.(2023)Zhao, Rao, Liu, Liu, Zhou, and Lu]{zhao2023unleashing}
Wenliang Zhao, Yongming Rao, Zuyan Liu, Benlin Liu, Jie Zhou, and Jiwen Lu.
\newblock Unleashing text-to-image diffusion models for visual perception.
\newblock In \emph{ICCV}, 2023.

\end{thebibliography}
}
\clearpage
\clearpage
\maketitlesupplementary

\tableofcontents

\appendix

\begin{figure}[t]
\centering
   \includegraphics[width=1\linewidth]{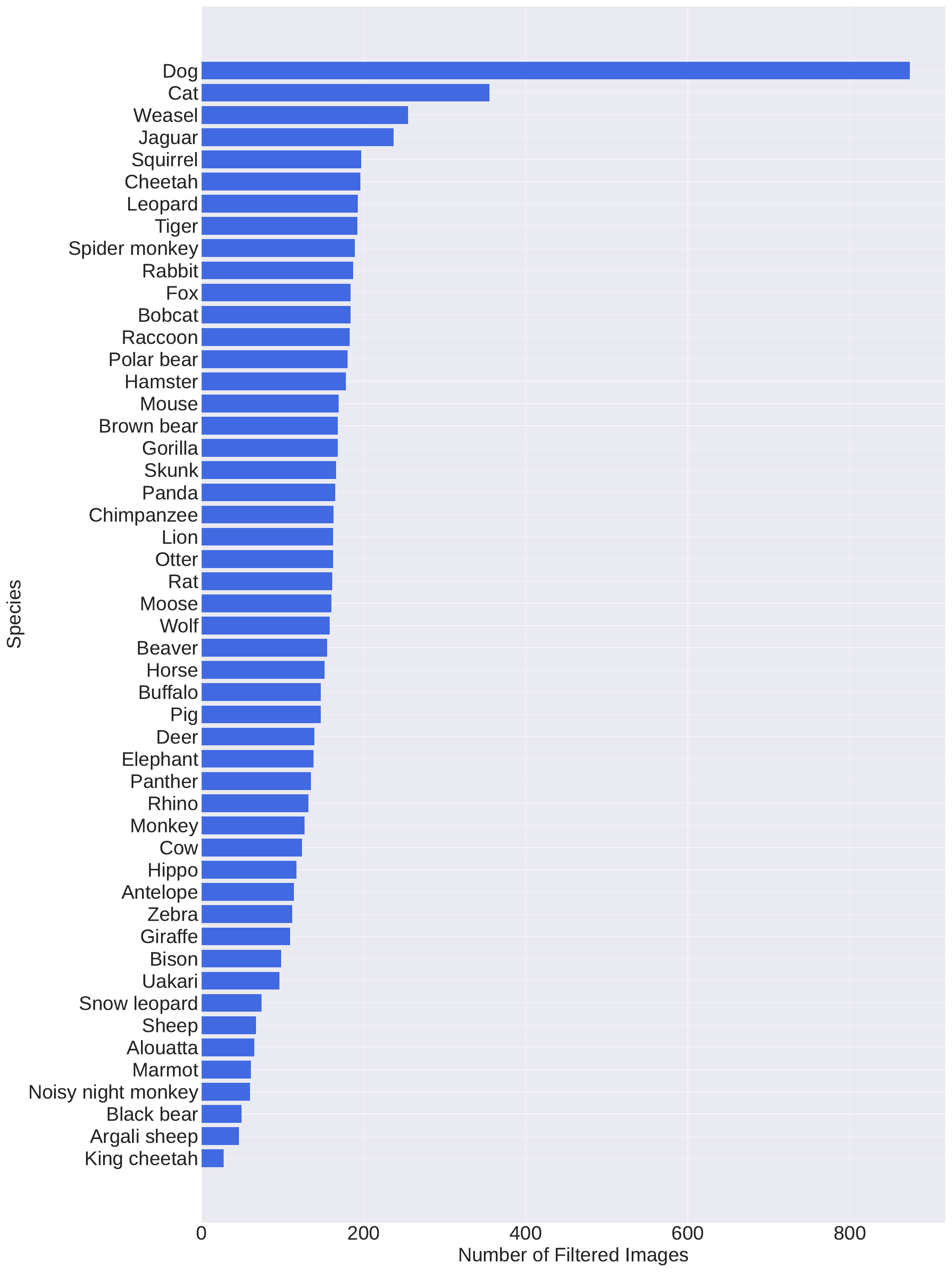}
   \caption{\textbf{Distribution of the filtered images across different species.} Note that only 50 species have annotated images.}
   \label{fig:8distribution_cleaned_json}
\end{figure}

\begin{figure*}[t]
\centering
   \includegraphics[width=0.95\linewidth]{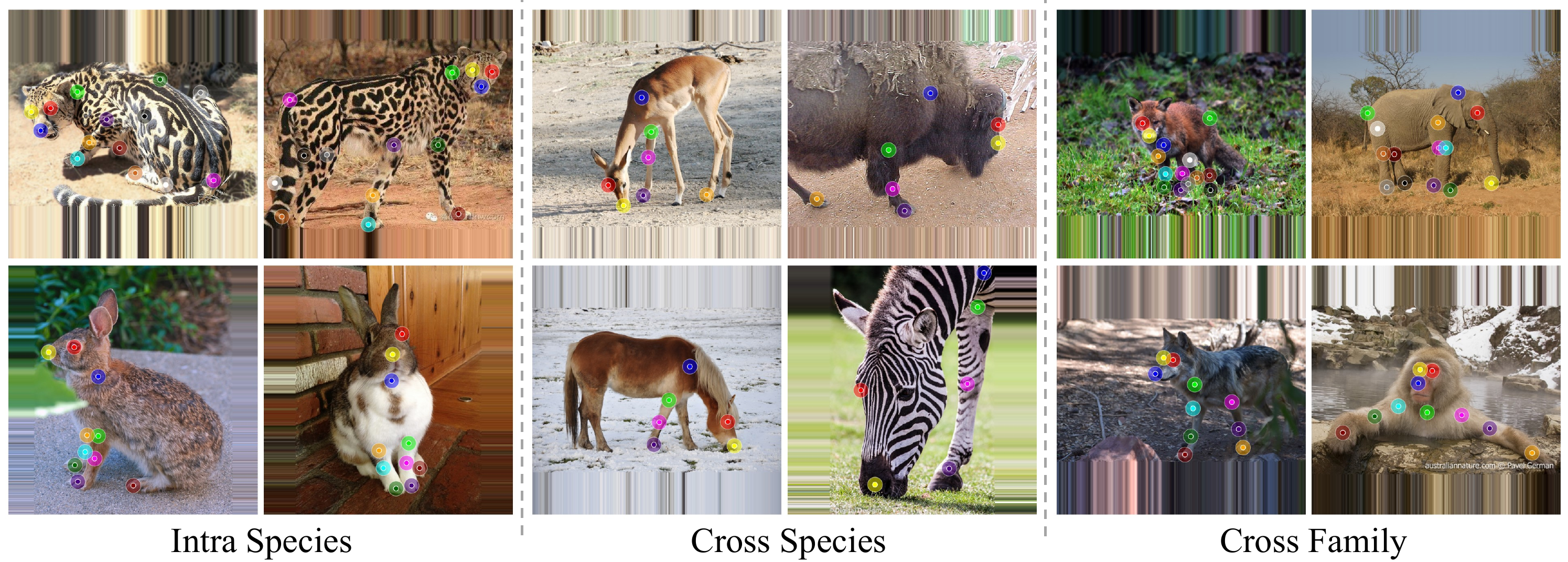}
   \vspace{-0.5em}
   \caption{\textbf{Sample image pairs of AP-10K benchmark} including intra species, cross species, and cross family.}
   \label{fig:8ap10k_sample}
   \vspace{-1em}
\end{figure*}

\section{Further Implementation Details}
\label{sec:more_implementation_details}
\myparagraph{Feature extraction.}
The extraction of SD and DINOv2 features is conducted in a manner similar to that described in Zhang \etal.~\cite{zhang2023tale}.
Specifically, the SD features are extracted from SD-1-5's UNet decoder layer 2, 5, and 8 at timestep 50 with an implicit captioner, and the DINOv2 features are extracted from the token facet of the 11th layer.

\myparagraph{Adaptive viewpoint alignment.}
For adaptive viewpoint (or pose) alignment in~\cref{sec:pose_alignment} in the main paper, we utilize segmentation masks from ODISE~\cite{xu2023open} to calculate the Instance Matching Distance (IMD). Considering the imbalanced viewpoint distribution in the images, ``horizontal flip" is employed as the primary viewpoint augmentation for all categories. Specifically for the bottle category, to accommodate its unique viewpoint variations, we further apply rotations of +90°, 180°, and -90° as additional augmented viewpoints.

\myparagraph{Pose-variant augmentation.}
In terms of pose-variant augmentation, we compute all the pair augmentations in a single batch and assign weights of 1 for both \textit{single-flip} and \textit{double-flip}, and a weight of 0.25 for the \textit{self-flip}. 
Note that pose-variant augmentation is not applied during training on the PF-Pascal dataset due to all image pairs in this dataset are of similar pose.

\myparagraph{Training.}
Our model is trained for 100k steps (equivalent to 2 epochs) on the SPair-71k dataset, and 250k steps on AP-10K (equivalent to 1 epoch) and PF-Pascal (equivalent to 85 epochs), with a mini-batch size of 1.
For a faster training, we pre-extract features from the visual foundation models offline and only train the post-processor online. This strategy significantly reduces the training duration, allowing it to be completed within just a few hours on a single GPU.

\section{Benchmarking AP-10K Dataset for\\ Semantic Correspondence}
\label{sec:benchmark}
\myparagraph{Image filtering.} To start with, we exclude images with fewer than three visible keypoints or with multiple instances of the target category, to make the dataset less ambiguous for semantic matching.

\myparagraph{Train/validation/test sets.} After the filtering, there exists an imbalance in the number of images per species within the AP-10K dataset, as illustrated in~\cref{fig:8distribution_cleaned_json}. 
To ensure a balanced evaluation across different species, we uniformly sample an equivalent number of images for validation and test sets across all species — specifically, $N_\text{val}=20$ for validation and $N_\text{test}=30$ for testing, in line with the protocol established by SPair-71k~\cite{min2019spair}. 
The remaining images constitute the training set. 
It is important to note that for these three species, king cheetah, argali sheep, and black bear, whose numbers of images after the filtering are below 50, we earmark these as a hold-out set without including them in the training set. Thereby, it can also provide a measure for evaluating the generalization capability of semantic correspondence methods. 

\myparagraph{Intra-species image pair sampling.} For each species, we construct all possible image matching pairs within each validation and test set (\ie,~$N_\text{val} \choose 2$ and $N_\text{test} \choose 2$) that are established in the previous step.
On the other hand, the training set exhibits a more significant variance in the number of images; to circumvent the unbalanced distribution that arises from quadratic pairing growth, we limit the pairing to a maximum of either $50 \times N_\text{train}$ or $N_\text{train} \choose 2$ pairs, whichever is fewer. 
Considering that the AP-10K dataset was not initially curated for the task of semantic correspondence, we apply an additional filtration criterion to the image pairs, retaining only those with a minimum of three mutual visible keypoints.
This results in a total number of 260,950 training, 8816 validation, and 20,630 testing image pairs.

\myparagraph{Cross-species and cross-family image pair sampling.} 
We also include correspondence matching pairs across different species and families.
For all 11 families with multiple species, we sample ${N_\text{val} \choose 1} \cdot {N_\text{val} \choose 1}$ validation pairs and ${N_\text{test} \choose 1} \cdot {N_\text{test} \choose 1}$ testing pairs for each family. For the cross-family setting, among all the ${21 \choose 2}$ combination of the total of 21 families, we only sample $N_\text{val}$ validation and ${N_\text{test}}$ testing pairs to save compute. A filtering process based on the mutually visible keypoints is also applied, yielding a total number of 4300 and 4200 validation pairs, alongside 9619 and 6300 testing pairs for cross-species and cross-family correspondence, respectively. Please refer to~\cref{fig:8ap10k_sample} for sample image pairs.

\begin{figure*}[t]
\centering
   \includegraphics[width=0.95\linewidth]{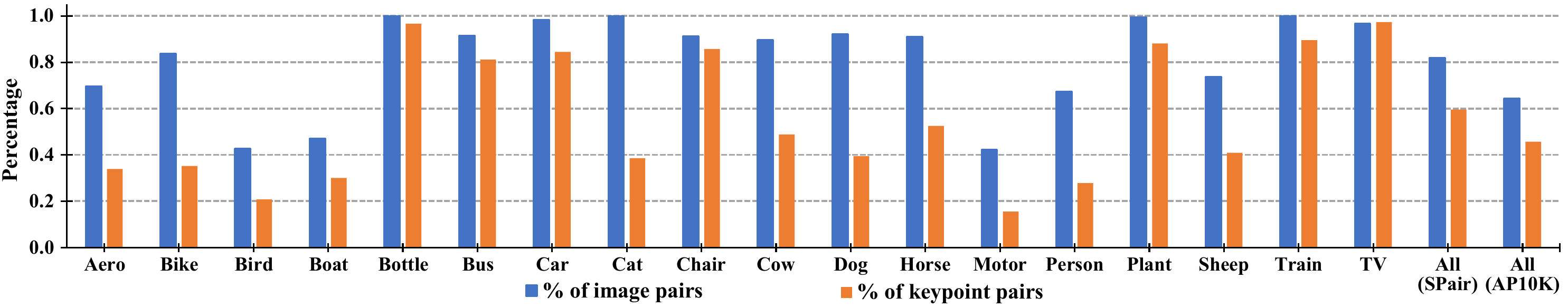}
   \caption{\textbf{Proportion of the geometry-aware subset with respect to image pair and keypoint pair.} We show the per-category results of SPair-71k as well as the average results of SPair-71k and AP-10K intra-species set.}
   \label{fig:8proportion}
\end{figure*}

\begin{table*}[t]
\centering
\small
\renewcommand{\arraystretch}{1.0}
\caption{\textbf{Semantically similar keypoint subgroups.} We list the keypoint subgroups for categories from both SPair-71k and AP-10K. The number in the bracket indicates the number of keypoints in each subgroup. The annotation in the index version will also be released.}
\label{tab:subgroups}
\begin{tabular}{l @{\hskip 3em} l @{\hskip 5em} c}
\toprule
 Dataset & Category & Subgroups \\
\midrule
\multirow{22}{*}{SPair-71k} 
& Aeroplane & $\mathcal{G}_\text{landing\_gear}$ (2), $\mathcal{G}_\text{engine\_front}$ (2), $\mathcal{G}_\text{wing\_end}$ (2), $\mathcal{G}_\text{engine\_back}$ (2), $\mathcal{G}_\text{wing\_foot\_front}$ (2), $\mathcal{G}_\text{wing\_foot\_back}$ (2), \\
&  & $\mathcal{G}_\text{tailplane\_end}$ (2), $\mathcal{G}_\text{tailplane\_foot\_front}$ (2), $\mathcal{G}_\text{tailplane\_foot\_back}$ (2) \\
\cmidrule(){2-3}

& Bicycle & $\mathcal{G}_\text{handle}$ (2), $\mathcal{G}_\text{seat\_back\_end}$ (2), $\mathcal{G}_\text{pedal}$ (2) \\
\cmidrule(){2-3}

& Bird & $\mathcal{G}_\text{wing\_end}$ (2), $\mathcal{G}_\text{foot}$ (2), $\mathcal{G}_\text{knee}$ (2), $\mathcal{G}_\text{hip}$ (2) \\
\cmidrule(){2-3}

& Boat & $\mathcal{G}_\text{upper\_front}$ (2), $\mathcal{G}_\text{upper\_side}$ (2), $\mathcal{G}_\text{upper\_back}$ (2),  $\mathcal{G}_\text{lower\_front}$ (2), $\mathcal{G}_\text{lower\_side}$ (2), $\mathcal{G}_\text{lower\_back}$ (2) \\
\cmidrule(){2-3}

& Bottle & $\mathcal{G}_\text{cap}$ (2), $\mathcal{G}_\text{neck}$ (2), $\mathcal{G}_\text{shoulder}$ (2), $\mathcal{G}_\text{body}$ (2), $\mathcal{G}_\text{base}$ (2) \\
\cmidrule(){2-3}

& Bus & $\mathcal{G}_\text{rearview\_mirror}$ (2), $\mathcal{G}_\text{light}$ (2), $\mathcal{G}_\text{licence\_plate}$ (2), $\mathcal{G}_\text{front\_fender}$ (4), $\mathcal{G}_\text{wheel}$ (4), $\mathcal{G}_\text{rear\_fender}$ (4), \\
&  & $\mathcal{G}_\text{window\_top\_corner}$ (4), $\mathcal{G}_\text{window\_bottom\_corner}$ (4) \\
\cmidrule(){2-3}

& Car & $\mathcal{G}_\text{rearview\_mirror}$ (2), $\mathcal{G}_\text{light}$ (2), $\mathcal{G}_\text{licence\_plate}$ (2), $\mathcal{G}_\text{brand\_logo}$ (2), $\mathcal{G}_\text{rear\_fender}$ (4), $\mathcal{G}_\text{wheel}$ (4), \\
&  & $\mathcal{G}_\text{front\_fender}$ (4), $\mathcal{G}_\text{window\_bottom\_corner}$ (4), $\mathcal{G}_\text{window\_top\_corner}$ (4) \\
\cmidrule(){2-3}

& Cat & $\mathcal{G}_\text{ear}$ (2), $\mathcal{G}_\text{paw}$ (4) \\
\cmidrule(){2-3}

& Chair & $\mathcal{G}_\text{cushion\_front}$ (2), $\mathcal{G}_\text{cushion\_back}$ (2), $\mathcal{G}_\text{leg}$ (4), $\mathcal{G}_\text{backrest\_top}$ (2), $\mathcal{G}_\text{armrest\_front}$ (2), $\mathcal{G}_\text{armrest\_back}$ (2) \\
\cmidrule(){2-3}

& Cow & $\mathcal{G}_\text{ear}$ (2), $\mathcal{G}_\text{hoof}$ (4), $\mathcal{G}_\text{knee}$ (4), $\mathcal{G}_\text{horn}$ (2) \\
\cmidrule(){2-3}

& Dog & $\mathcal{G}_\text{ear}$ (2), $\mathcal{G}_\text{paw}$ (4) \\
\cmidrule(){2-3}

& Horse & $\mathcal{G}_\text{ear}$ (2), $\mathcal{G}_\text{hoof}$ (4), $\mathcal{G}_\text{knee}$ (4) \\
\cmidrule(){2-3}

& Motorbike & $\mathcal{G}_\text{rearview\_mirror}$ (2), $\mathcal{G}_\text{handle}$ (2) \\
\cmidrule(){2-3}

& Person & $\mathcal{G}_\text{shoulder}$ (2), $\mathcal{G}_\text{elbow}$ (2), $\mathcal{G}_\text{wrist}$ (2), $\mathcal{G}_\text{knee}$ (2), $\mathcal{G}_\text{ankle}$ (2), $\mathcal{G}_\text{foot}$ (2) \\
\cmidrule(){2-3}

& Pottedplant & $\mathcal{G}_\text{top}$ (4), $\mathcal{G}_\text{side\_wall}$ (2), $\mathcal{G}_\text{bottom}$ (2) \\
\cmidrule(){2-3}

& Sheep & $\mathcal{G}_\text{ear}$ (2), $\mathcal{G}_\text{hoof}$ (4), $\mathcal{G}_\text{knee}$ (4), $\mathcal{G}_\text{horn}$ (2) \\
\cmidrule(){2-3}

& Train & $\mathcal{G}_\text{front\_top}$ (2), $\mathcal{G}_\text{front\_bottom}$ (2), $\mathcal{G}_\text{back\_top}$ (2), $\mathcal{G}_\text{back\_bottom}$ (2), $\mathcal{G}_\text{window\_top\_outer\_corner}$ (2),  \\
&  & $\mathcal{G}_\text{window\_bottom\_outer\_corner}$ (2), $\mathcal{G}_\text{window\_top\_inner\_corner}$ (2), $\mathcal{G}_\text{window\_bottom\_inner\_corner}$ (2), $\mathcal{G}_\text{front\_light}$ (2) \\
\cmidrule(){2-3}

& Tvmonitor & $\mathcal{G}_\text{outer\_corner}$ (4), $\mathcal{G}_\text{outer\_side}$ (4), $\mathcal{G}_\text{inner\_corner}$ (4), $\mathcal{G}_\text{inner\_side}$ (4) \\
\midrule

AP-10K &All& $\mathcal{G}_\text{shoulder}$ (2), $\mathcal{G}_\text{foot}$ (4), $\mathcal{G}_\text{knee}$ (4), $\mathcal{G}_\text{hip}$ (2) \\
\bottomrule
\end{tabular}
\end{table*}

\begin{figure*}[t]
  \centering
 \begin{subfigure}{1.0\linewidth}
 \centering
 \includegraphics[width=0.985\linewidth]{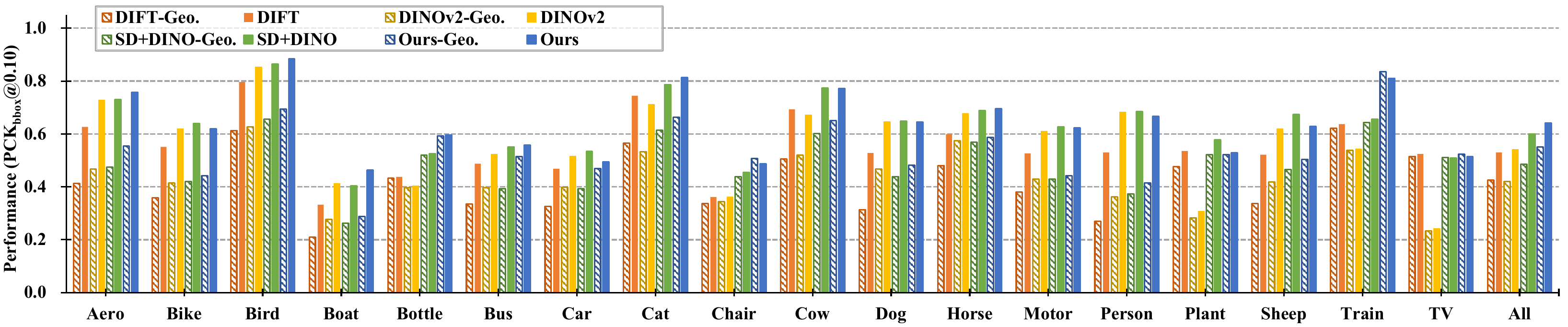}
     \caption{\textbf{Performance of the unsupervised methods.}}
   \label{fig:8unsuper_performance}
  \end{subfigure}
  \par\smallskip
  \begin{subfigure}{1.0\linewidth}
    \includegraphics[width=1\linewidth]{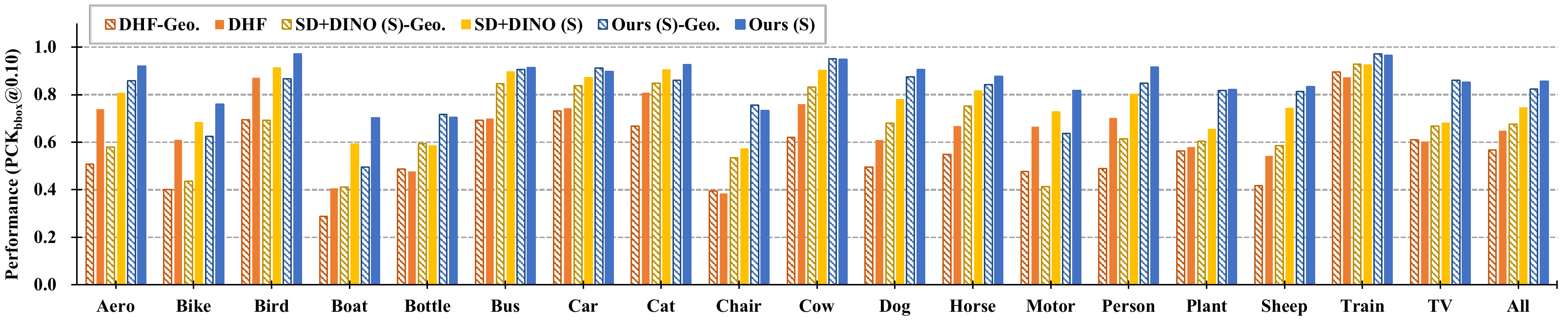}
     \caption{\textbf{Performance of the supervised methods.}}
   \label{fig:8super_performance}
  \end{subfigure}
  \caption{\textbf{Per-category performance of the state-of-the-art methods and ours (\blue{blue}).} We report both the geometry-aware subset (Geo.) and the standard set on SPair-71k. Our methods consistently outperform previous arts across all categories.}
  \label{fig:8performance}
  \vspace{-0.5em}
\end{figure*}

\begin{figure*}[t]
\centering
   \includegraphics[width=0.975\linewidth]{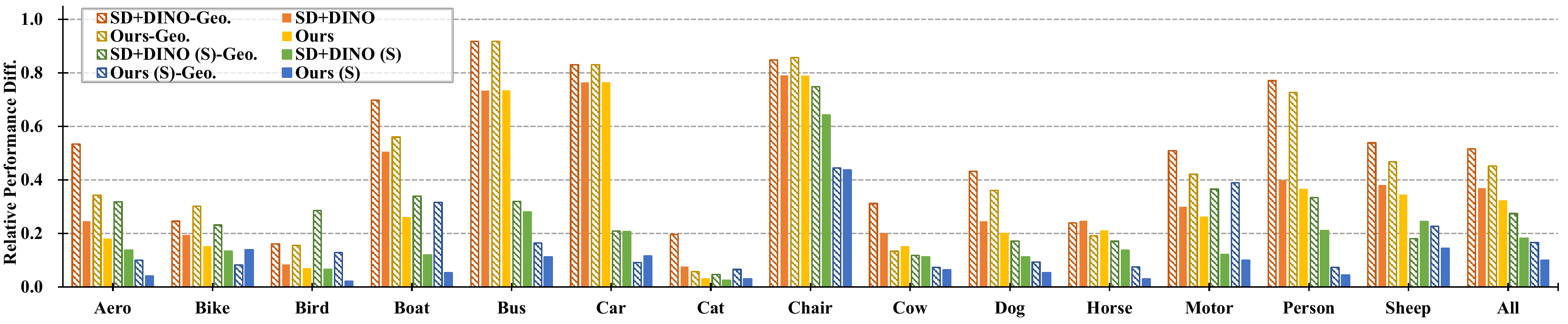}
   \vspace{-0.5em}
   \caption{\textbf{Per-category evaluation of the sensitivity to pose variations.} Both our zero-shot (\yellow{yellow}) and supervised methods (\blue{blue}) considerably improve the robustness to pose variations on both the geometry-aware set (Geo., hashed bar) and the standard set (solid bar) compared to the state-of-the-art methods~\cite{zhang2023tale}. We exclude categories that only have one azimuth-variation subset.}
   \label{fig:8sensitivity}
   \vspace{-0.5em}
\end{figure*}

\section{Details on Geo-Aware Correspondence}
\label{sec:geo-aware-details}

\myparagraph{Keypoint subgroups.}
We list the keypoint subgroups of each category in~\cref{tab:subgroups}.
We exclude very few parts (nostril, eyes, \etc) that are close to each other and thus cannot be easily distinguished by existing metrics.
We suggest that an improved metric (\eg, a keypoint can be only regarded as a prediction to its nearest ground truth point) can make up this issue.

\myparagraph{Per-category proportion.}
We show the average proportion of the geometry-aware subset with respect to both image pairs and keypoint pairs for each category in~\cref{fig:8proportion}. 
For most of the categories, the geometry-aware subset accounts for a considerable fraction of all pairs.

Notably, due to the unbalanced pose distribution exhibited in specific categories of the SPair-71k (\eg~bottles, potted plants, TVs, and trains) where image pairs often share similar poses, almost all keypoint subgroups in these categories are mutually visible, which results in proportions to be near 100\%. 
In contrast, the AP-10K dataset, comprised solely of animal images, does not exhibit this imbalance.

\myparagraph{Per-category performance.}
In \cref{fig:8unsuper_performance} and \cref{fig:8super_performance}, we provide detailed per-category performance for both unsupervised and supervised state-of-the-art methods on the geometry-aware subset and the standard set. These figures provide an expanded view of~\cref{fig:3.2per-category} from the main paper. 
Regardless of the method or category, performance on the geometry-aware subset consistently lags behind that of the standard set.

Additionally, in~\cref{fig:8sensitivity}, we offer a per-category analysis of pose variation sensitivity. 
The results for both unsupervised and supervised variants of SD+DINO~\cite{zhang2023tale} are presented, comparing their performance on both the geometry-aware and standard sets. 
This analysis serves as an extended version of~\cref{fig:3.3per-category-az} from the main paper. 
The findings clearly show that sensitivity to pose variation is considerably higher in the geometry-aware subset across all categories and methodologies.

\section{Additional Analysis}

\subsection{Detailed Performance on Geo-Aware Subset}

We provide the per-category performance on the geometry-aware subset in~\cref{fig:8performance} as well as the pose-sensitivity analysis of our methods in \cref{fig:8sensitivity}.

\subsection{Detailed Analysis on Window Soft Argmax}

\begin{figure}[t]
\centering
   \includegraphics[width=1\linewidth]{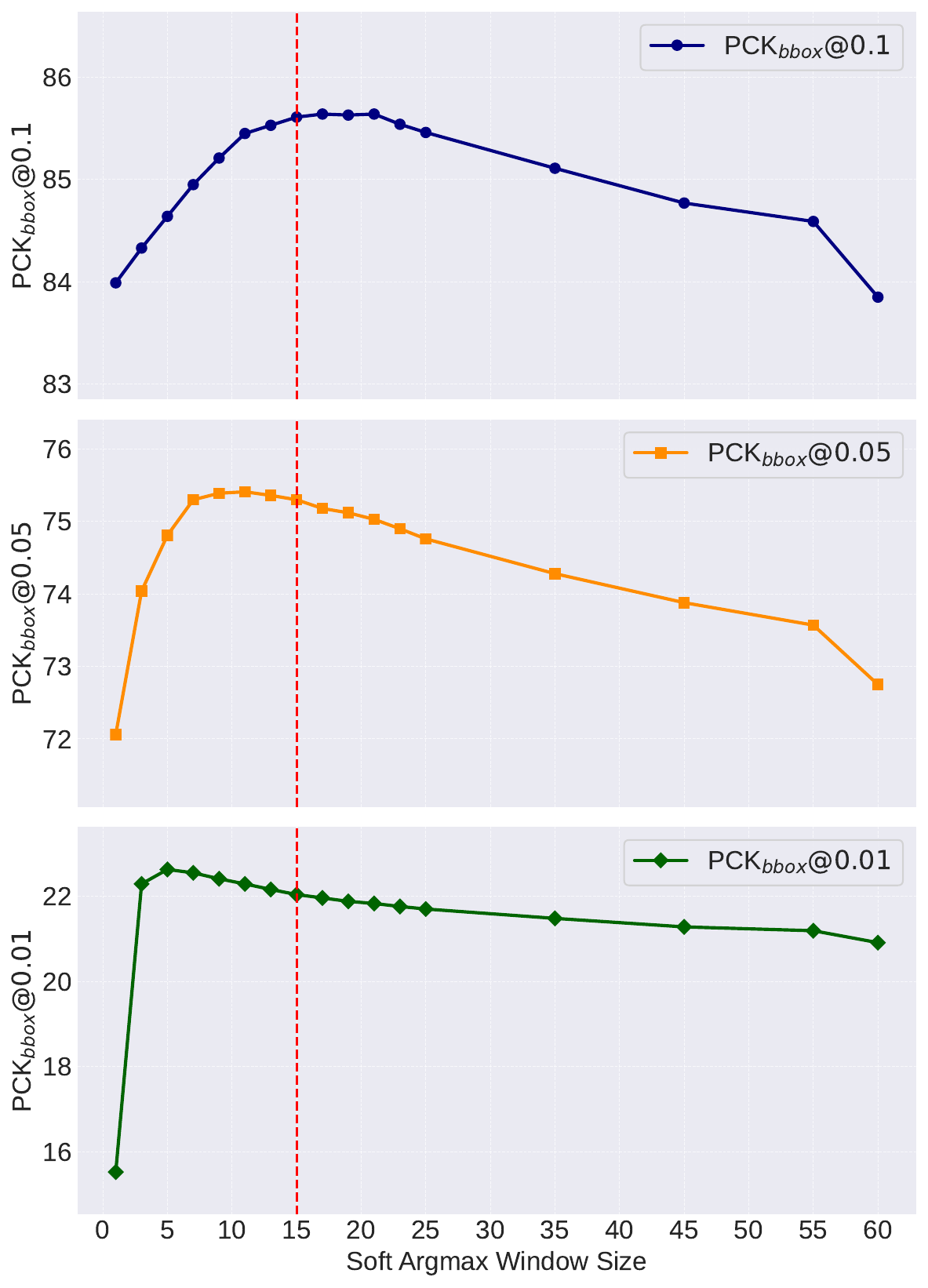}
   \caption{\textbf{Performance of different PCK levels vs. soft argmax window size.} We test the performance on the SPair-71k dataset and set the window size as 15 for optimal balance.}
   \label{fig:8window_size}
\end{figure}

\myparagraph{Performance in accordance with window size.} We evaluate the effect of soft-argmax's window size on the performance at different PCK thresholds. As depicted in~\cref{fig:8window_size}, the performance across all PCK levels initially improves and then declines as the window size increases from 0 (hard argmax) to 60 (soft argmax). Notably, the peak PCK values for 0.01, 0.05, and 0.1 are observed at window sizes of 5, 11, and 17, respectively. We opt for a window size of 15 to achieve an optimal balance in performance.

\myparagraph{Comparison with Gaussian kernel soft argmax.}
Previous work~\cite{lee2019sfnet} also explored a trade-off solution between hard and soft argmax by applying a Gaussian kernel on the feature map, centered at the hard argmax position.

We also search different $\sigma$ values for the Gaussian kernel to achieve the best performance across different PCK levels.
We then compare our window soft argmax with the kernel soft argmax in~\cref{tab:compare_kernel} on different peak PCK levels and the default value as reported in~\cite{lee2019sfnet}.
Our window soft argmax consistently outperforms kernel argmax across all settings, suggesting the superiority of our approach.
We hypothesize that this is because when using the argmax-centered Gaussian kernel to scale the similarity map, it makes the similarity map biased to argmax locations, while our method treats the window region with the same scale.

\begin{table}[t]
\centering
\renewcommand{\arraystretch}{1.1}
\caption{\textbf{Comparison with Gaussian kernel soft argmax on SPair-71k.} Default and peak values for each PCK level are reported for both methods, with the best results \textbf{bolded}.}
\label{tab:compare_kernel}
\resizebox{0.475\textwidth}{!}{
\begin{tabular}{llccc}
\toprule
 Setting &{Method} & PCK@0.01 & PCK@0.05 & PCK@0.10 \\
\midrule
\multirow{2}{*}{Default}
&Kernel&19.7& 73.5& 84.3\\
&Window&{22.0}& {75.3}& {85.6}\\
\hline
\multirow{2}{*}{Best PCK@0.01}
&Kernel&22.4& 75.0& 84.9\\
&Window&\textbf{22.6}& {75.3}& {85.0}\\
\hline
\multirow{2}{*}{Best PCK@0.05}
&Kernel&22.3& 75.3& 85.3\\
&Window&{22.3}& \textbf{75.4}& {85.5}\\
\hline
\multirow{2}{*}{Best PCK@0.10}
&Kernel&21.9& 75.1& 85.5\\
&Window&{22.0}& {75.2}& \textbf{85.7}\\
\bottomrule
\end{tabular}
}
\end{table}

\myparagraph{Training with window soft argmax.}
We also experiment if applying the window soft argmax during training is beneficial.
As shown in~\cref{tab:window_training}, applying window soft argmax during training hurts the PCK performances with the loose thresholds, while helping the stricter threshold (\ie, PCK@0.01).
Our hypothesis is that applying windows during training helps the model focus on the local region but overlook global information.

\begin{table}[t]
\centering
\renewcommand{\arraystretch}{1.1}
\caption{\textbf{Effect of applying window soft argmax during training.} We train all the post-processors on SPair-71k for one epoch and from scratch. The best results are \textbf{bolded}.}
\label{tab:window_training}
\resizebox{0.475\textwidth}{!}{
\begin{tabular}{lccc}
\toprule
 Setting  & PCK@0.01 & PCK@0.05 & PCK@0.10 \\
\midrule
Window Soft Argmax (7)&21.4& 69.0& 79.0\\
Window Soft Argmax (15)&21.8& 69.5& 80.1\\
Window Soft Argmax (22)&\textbf{22.1}& 70.3& 81.2\\
\hline
Soft Argmax&20.4& \textbf{70.8}& \textbf{82.1}\\
\bottomrule
\end{tabular}
}
\end{table}

\begin{table*}[h]
    \centering
    \setlength{\tabcolsep}{1.1mm}{
    \renewcommand{\arraystretch}{1.05}
    \caption{\textbf{Leave-one-out ablation study on SPair-71k.} We report the \textit{per image} results and four metrics introduced in~\cite{aygun2022demystifying} (\ie, Jitter, Miss, Swap, and PCK$^\dag$) for a detailed analysis of the effect of each module. The best results are \textbf{bold}.}
\label{tab:leave-one-out}
        \begin{tabular}{lccccc|ccc}
\toprule
Variations & Jitter$\downarrow$ & Miss$\downarrow$ & Swap$\downarrow$ & $\textrm{Swap}^{LR}$$\downarrow$ &  PCK$^\dag$@0.1$\uparrow$ & PCK@0.01$\uparrow$ & PCK@0.05$\uparrow$ & PCK@0.1$\uparrow$ \\
\midrule
SD+DINO (S) \textbf{(Baseline)} & 9.7 & 13.7 & 15.8 & 9.4 & 70.5& 9.6& 57.7& 74.6 \\
\midrule
w/o Dense Training Objective & 8.3  & 11.8 & 13.7 & 8.5  & 74.5 & 15.2& 64.5& 78.3 \\
w/o Pose-variant Augmentation &7.4  & 10.0 & 13.9 & 8.7  & 76.1 & 19.0& 70.3 & 81.5\\
w/o Perturbation \& Dropout & 6.9  &9.9 & 12.3 & 7.2  & 77.8 & 20.3& 71.8 & 82.3\\
w/o Window Soft Argmax & 8.1  & 9.8 & 14.1 & 8.7  & 76.1 & 15.1& 69.3 & 81.3\\
\hline
\textbf{Ours} &6.9  & 9.3 & 12.0 & 7.0  & 78.7&21.6& 72.6 & 82.9\\
\textbf{Ours w/ AP-10k Pretraining} & \textbf{6.1}  & \textbf{8.7} &\textbf{10.4}  & \textbf{5.6}  & \textbf{80.9}& \textbf{22.0}& \textbf{75.3} & \textbf{85.6}\\
\bottomrule
            \end{tabular}
    }

\end{table*}

\subsection{Discussion on Generalizability}

As shown in the main paper, we validate the generalizability of our method by training on AP-10K intra-species set and testing on cross-species and cross-family subsets. Here, we extend this analysis with additional tests:

\myparagraph{Training on PF-PASCAL and testing on other datasets.} 
We evaluate the generalizability of our method by training it on PF-PASCAL and then testing it on SPair-71k and AP-10K intra-species test sets (see~\cref{tab:8generalize_pascal}). 
While previous studies~\cite{cho2022cats++, huang2022learning} have noted a potential performance decrease due to models' overfitting to the limited distribution of pose variation in PF-PASCAL,
our method consistently outperforms across different datasets and PCK thresholds, demonstrating its robustness.

\begin{table}[t]
\centering
\renewcommand{\arraystretch}{1.025}
\caption{\textbf{Generalizability test with training on PF-PASCAL.} We test the generalizability of our method by training the model on the PF-PASCAL dataset and testing on the SPair-71k and AP-10K intra-species (I.S.) test set. The best results are \textbf{bold}.}
\label{tab:8generalize_pascal}

\resizebox{0.475\textwidth}{!}{
\begin{tabular}{lcccccc}
\toprule
&\multicolumn{3}{c}{SPair-71k} &\multicolumn{3}{c}{AP-10K-I.S.}\\
\cmidrule(l){2-4} \cmidrule(l){5-7}
  {Method} & 0.01 & 0.05 & 0.10 & 0.01 & 0.05 & 0.10\\
\midrule

SCorrSAN~\cite{huang2022learning} & 1.5& 18.4 & 32.7 & - & - & - \\
CATs++~\cite{cho2022cats++} &2.1 & 19.7 & 32.0 & - & - & - \\
DHF~\cite{luo2023diffusion} &4.6 & 30.1 & 41.8 & 7.3 & 37.0 & 49.1 \\
SD+DINO (S)~\cite{zhang2023tale} &\textbf{5.3} & 34.1 & 46.9 & 8.2 & 43.4 & 59.2 \\
\textbf{Ours}&\textbf{5.3} & \textbf{37.1} & \textbf{54.3} & \textbf{10.1} & \textbf{44.0} & \textbf{62.5}\\
\bottomrule
\end{tabular}
}
\end{table}

\myparagraph{Training on SPair-71k and testing on AP-10K and PF-PASCAL.}
In a similar vein, we trained our model on the SPair-71k dataset and evaluated its performance on PF-PASCAL and AP-10K intra-species test sets (see~\cref{tab:8generalize_spair}). 
The findings mirrored those from~\cref{tab:8generalize_pascal}, with our approach achieving the best results across all datasets and PCK metrics, confirming its generalizability again.

\begin{table}[t]
\centering
\renewcommand{\arraystretch}{1.025}
\caption{\textbf{Generalizability test with training on SPair-71k.} We test the generalizability of our method by training the model on the SPair-71k dataset and testing on the PF-PASCAL and AP-10K intra-species (I.S.) test set. The best results are \textbf{bold}.}
\label{tab:8generalize_spair}
\resizebox{0.475\textwidth}{!}{
\begin{tabular}{lcccccc}
\toprule
&\multicolumn{3}{c}{PF-PASCAL} &\multicolumn{3}{c}{AP-10K-I.S.}\\
\cmidrule(l){2-4} \cmidrule(l){5-7}
  {Method} & 0.05 & 0.10 & 0.15 & 0.01 & 0.05 & 0.10\\
\midrule
SCorrSAN~\cite{huang2022learning} &54.5 & 71.2 & 78.8 & - & - & - \\
CATs++~\cite{cho2022cats++} &54.8 & 68.7 & 76.1 & - & - & - \\
DHF~\cite{luo2023diffusion} &64.2 & 77.8 & 84.0 & 9.3 & 42.0 & 55.2 \\
SD+DINO (S)~\cite{zhang2023tale} &68.9 & 81.7 & 87.2 & 9.7 & 50.4 & 65.9 \\
\textbf{Ours}&\textbf{74.0} & \textbf{85.3} & \textbf{89.7} & \textbf{16.5} & \textbf{56.7} & \textbf{70.2}\\
\bottomrule
\end{tabular}
}
\vspace{-0.5em}
\end{table}

\subsection{Additional Ablation Analysis under Supervised Setting}
\label{sec:leave-one-out}

To further evaluate the effect of each component on improving semantic correspondence, we conduct a leave-one-out ablation analysis. 
For an in-depth understanding of the specific improvements, we incorporate the breakdown analysis protocol from "Demystifying"~\cite{aygun2022demystifying} into our ablation study. 
This analysis introduces four metrics, as delineated in~\cite{aygun2022demystifying}: 
1) \textit{Jitter}: the ratio of matches near their correct locations; 
2) \textit{Miss}: the ratio of points incorrectly matched to the background; 
3) \textit{Swap}: the ratio of matches that are in the correct area but nearer to a different semantic part; 
4) \textit{PCK$^\dag$}: the PCK metric adjusted to exclude \textit{Swap} errors. 
For comprehensive details, please see Sec. 4.1 of~\cite{aygun2022demystifying}. 
To advance our evaluation of geometry-aware correspondence further, we introduce an additional metric, \textit{Swap$^{LR}$}, for geometric confusion (left/right) cases.

As shown in~\cref{tab:leave-one-out}, our method significantly improves \textit{Jitter}, \textit{Miss}, \textit{Swap}, \textit{Swap$^{LR}$} by 37.1\%, 36.5\%, 36.0\%, and 40.4\%, respectively. 
Specifically, the integration of spatial context through our proposed dense training objective and the window soft argmax technique notably boosts the performance for \textit{Jitter}, \textit{Swap}, and \textit{Swap$^{LR}$}, which relies on detailed spatial understanding. 
Besides, the dense training objective also contributes largely in overcoming the \textit{Miss} error, we hypothesize that the soft argmax operator in dense training objective can effectively suppress the background noise.
Moreover, by encouraging the pose-awareness, the proposed pose-variant pair augmentation notably reduces both the \textit{Swap} errors, and especially the geomety-aware \textit{Swap$^{LR}$} error.

In summary, the improvement in \textit{Swap$^{LR}$} metric further validates the effectiveness of our designs in improving the geometric-awareness of the pretrained features, while the gain in \textit{Miss} showcases that our method also reduces mismatches to the image background.

\subsection{Ablation Study under Unsupervised Setting}
In the main paper, our zero shot method consists of two techniques: adaptive pose alignment and window soft argmax.
In this section, we further ablate different techniques to evaluate the effectiveness of each module under the unsupervised setting.

\begin{table*}[t]
\centering
\small
\renewcommand{\arraystretch}{1.0}
\caption{\textbf{Ablation study under unsupervised setting.} We report the PCK@$\alpha_\text{bbox}$ results on both the standard set (Std.) and geometry-aware set (Geo.) of SPair-71k. The best performances are \textbf{bold}.}
\label{tab:8window_zero-shot}
\begin{tabular}{ll @{\hskip 2em} cccccc}
\toprule
&&\multicolumn{3}{c}{SPair-71k (Std.)} &\multicolumn{3}{c}{SPair-71k (Geo.)}\\
\cmidrule(l){3-5} \cmidrule(l){6-8}
  Method Variants& {Inference Strategy} & 0.01 & 0.05 & 0.10 & 0.01 & 0.05 & 0.10\\
\midrule
\multirow{5}{*}{SD+DINO~\cite{zhang2023tale}}
&Argmax Inference (Default) &7.9 & 44.7 & 59.9 & 5.3 & 34.5 & 49.3 \\
&Soft Argmax Inference &6.4 & 36.5 & 53.7 & 6.4 & 36.5 & 53.7 \\
\cmidrule(lr){2-8}
&Window Soft Argmax (3)&\textbf{10.0} & 45.9 & 60.1 & \textbf{6.7} & 35.5 & 49.6 \\
&Window Soft Argmax (5)&9.9 & \textbf{46.3} & 60.5 & 6.6 & \textbf{35.8} & 50.1 \\
&Window Soft Argmax (11)&8.7 & 45.3 & \textbf{61.3} & 5.5 & 34.3 & \textbf{51.1}\\
\midrule
\multirow{5}{*}{SD+DINO~\cite{zhang2023tale}
\textbf{ w/ Adapt. Pose}}
&Argmax Inference (Default) &8.9 & 48.7 & 64.2 & 6.3 & 39.6 & 55.0 \\
&Soft Argmax Inference &7.6 & 40.7 & 58.4 & 4.1 & 29.0 & 48.2 \\
\cmidrule(lr){2-8}
&Window Soft Argmax (3)&\textbf{11.2} & 49.7 & 64.3 & \textbf{8.3} & 40.8 & 55.4 \\
&Window Soft Argmax (5)&11.1 & \textbf{50.1} & 64.8 & 8.1 & \textbf{41.1} & 56.0 \\
&Window Soft Argmax (11)&9.9 & 49.1 & \textbf{65.4} & 6.9 & 39.5 & \textbf{56.8}\\
\bottomrule
\end{tabular}
\end{table*}

As shown in~\cref{tab:8window_zero-shot}, our adaptive pose alignment technique consistently improves the performance across all five inference settings. Additionally, under both with or without adaptive pose alignment settings, our window soft argmax method consistently boosts the performance on both the geometry-aware subset and standard set, outperforming either the argmax or soft argmax.
This further demonstrates the effectiveness of our method.

\section{Additional Results}

\subsection{Alternative Metrics for Pose Alignment}
\label{sec:alternative_pose_alignment}
\begin{table}[t]
\centering
\renewcommand{\arraystretch}{1.1}
\caption{\textbf{Effect of different adaptive pose alignment metric.} Alternative approaches with relaxed conditions can achieve very competitive results that are much better than the baseline.}
\label{tab:alternative_pose_alignment}
\resizebox{0.475\textwidth}{!}{
\begin{tabular}{lccc}
\toprule
 Setting  & PCK@0.01 & PCK@0.05 & PCK@0.10 \\
\midrule
None \textbf{(Baseline)}&7.9& 44.7& 59.9\\
\hline
Mutual-NN&8.5& 47.8& 63.1\\
SAM-mask~\cite{kirillov2023segment}&8.6& 48.5& 64.0\\
ODISE-mask~\cite{xu2023open} \textbf{(Default)}&{8.9}& {48.7}& {64.2}\\
\bottomrule
\end{tabular}
}
\vspace{-0.75em}
\end{table}

In our adaptive pose alignment method, we leverage the mask of the source image, obtained through an off-the-shelf segmentation method, ODISE~\cite{xu2023open}, to calculate the matching distance. 
While this mask is solely used for pose alignment and does not restrict the solution space for the target image, we propose more flexible approaches to calculate the metric for pose alignment.

Firstly, as an alternative to generating masks based on object categories (as in ODISE), we can employ a query-point-based segmentation method, \eg, SAM~\cite{kirillov2023segment}, to obtain the instance mask. 
Such setting has a more relaxed condition because the semantic correspondence task naturally provides query keypoints of the instance in the source image.
Furthermore, we can eliminate the need for masks at all by using the average distance of mutual nearest-neighbor pixels as the alignment metric. 
As shown in the~\cref{tab:alternative_pose_alignment}, both alternative metrics yield highly competitive results, significantly surpassing our baseline.

\subsection{Qualitative Results on AP-10K}
\label{sec:qualitative_ap10k}
\begin{figure*}[t]
\centering
   \includegraphics[width=0.95\linewidth]{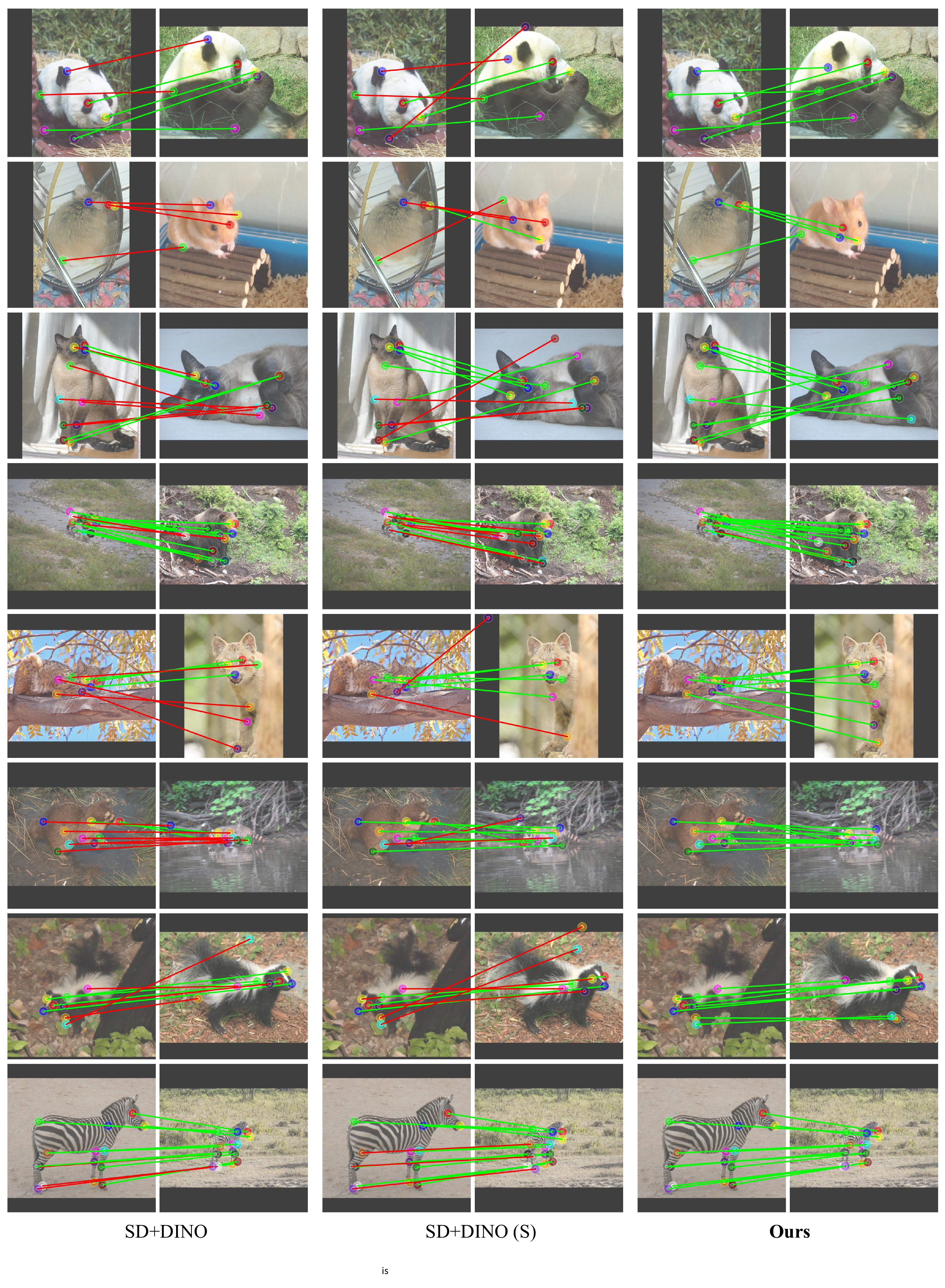}
   \caption{\textbf{Qualitative comparison on the AP-10K intra-species set.}}
   \label{fig:8qual_results_intra_species.pdf}
\end{figure*}

\begin{figure*}[t]
\centering
   \includegraphics[width=0.95\linewidth]{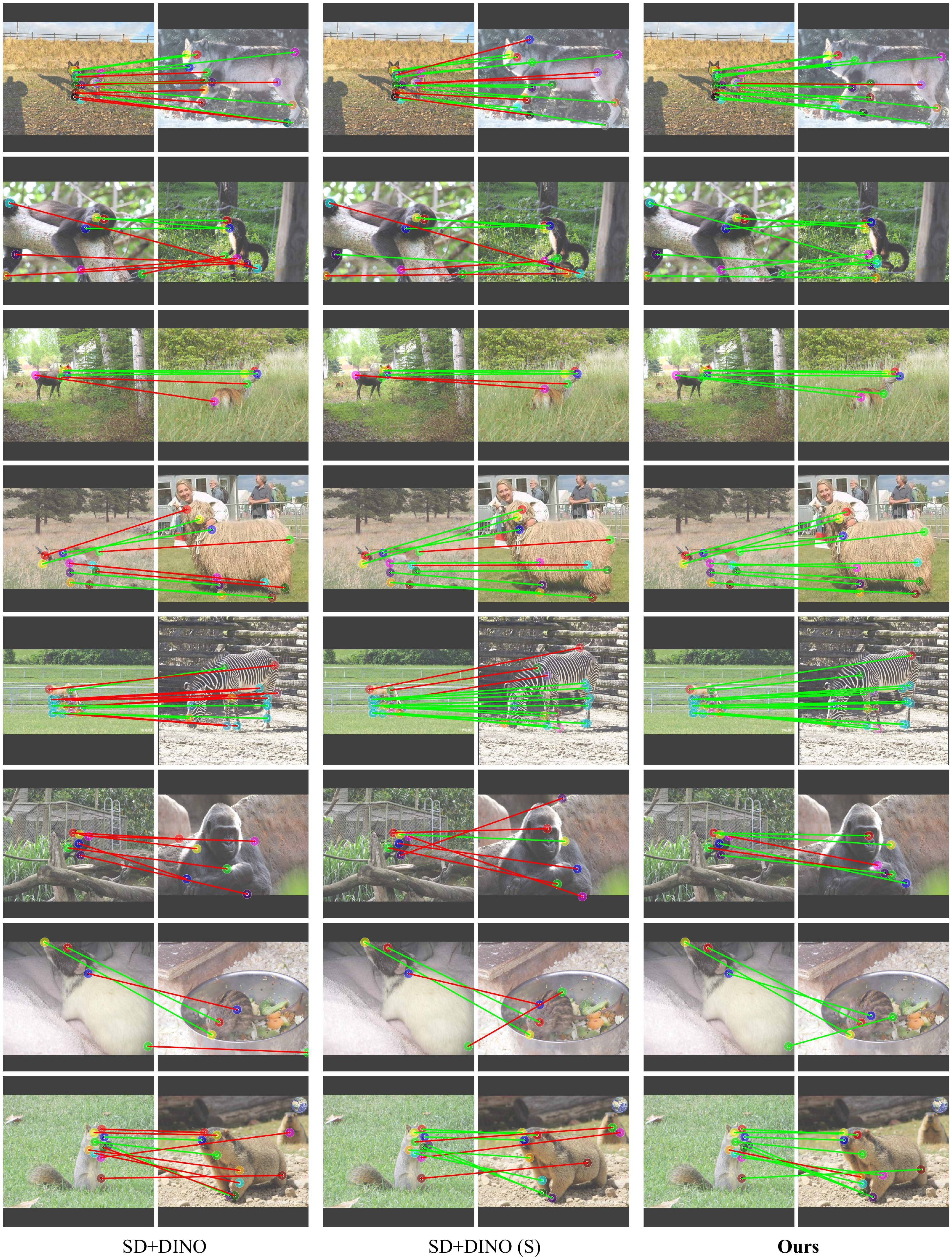}
   \caption{\textbf{Qualitative comparison on the AP-10K cross-species set.}}
   \label{fig:8qual_results_cross_species.pdf}
\end{figure*}

\begin{figure*}[t]
\centering
   \includegraphics[width=0.95\linewidth]{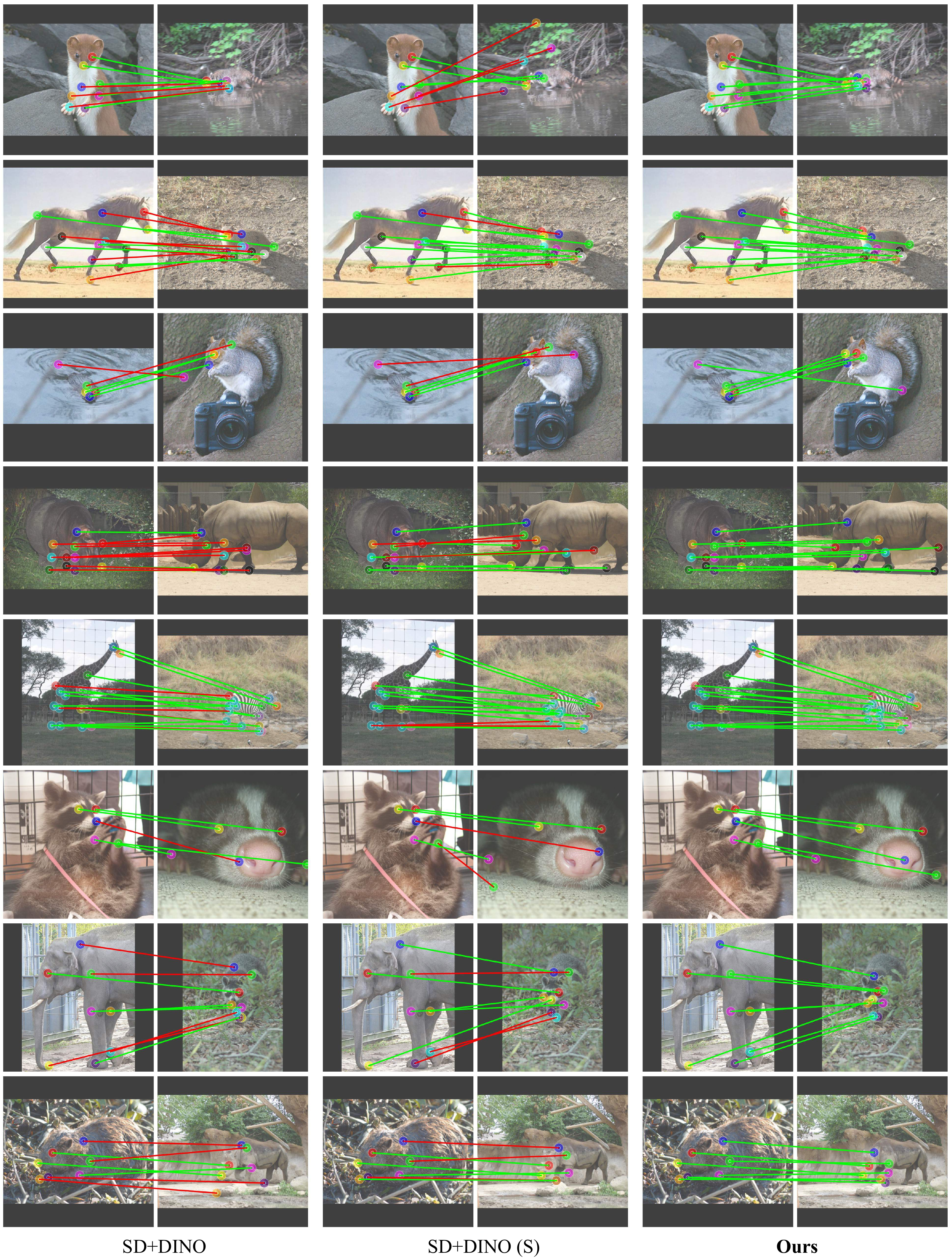}
   \caption{\textbf{Qualitative comparison on the AP-10K cross-family set.}}
   \label{fig:8qual_results_cross_families.pdf}
\end{figure*}

We show the qualitative comparison of our supervised methods with both unsupervised and supervised versions of SD+DINO~\cite{zhang2023tale} on AP-10K intra-species (\cref{fig:8qual_results_intra_species.pdf}), cross-species (\cref{fig:8qual_results_cross_species.pdf}), and cross-family (\cref{fig:8qual_results_cross_families.pdf}) subset.

\subsection{Additional Qualitative Results on SPair-71k}
\label{sec:qualitative_spair}
\begin{figure*}[t]
\centering
   \includegraphics[width=0.95\linewidth]{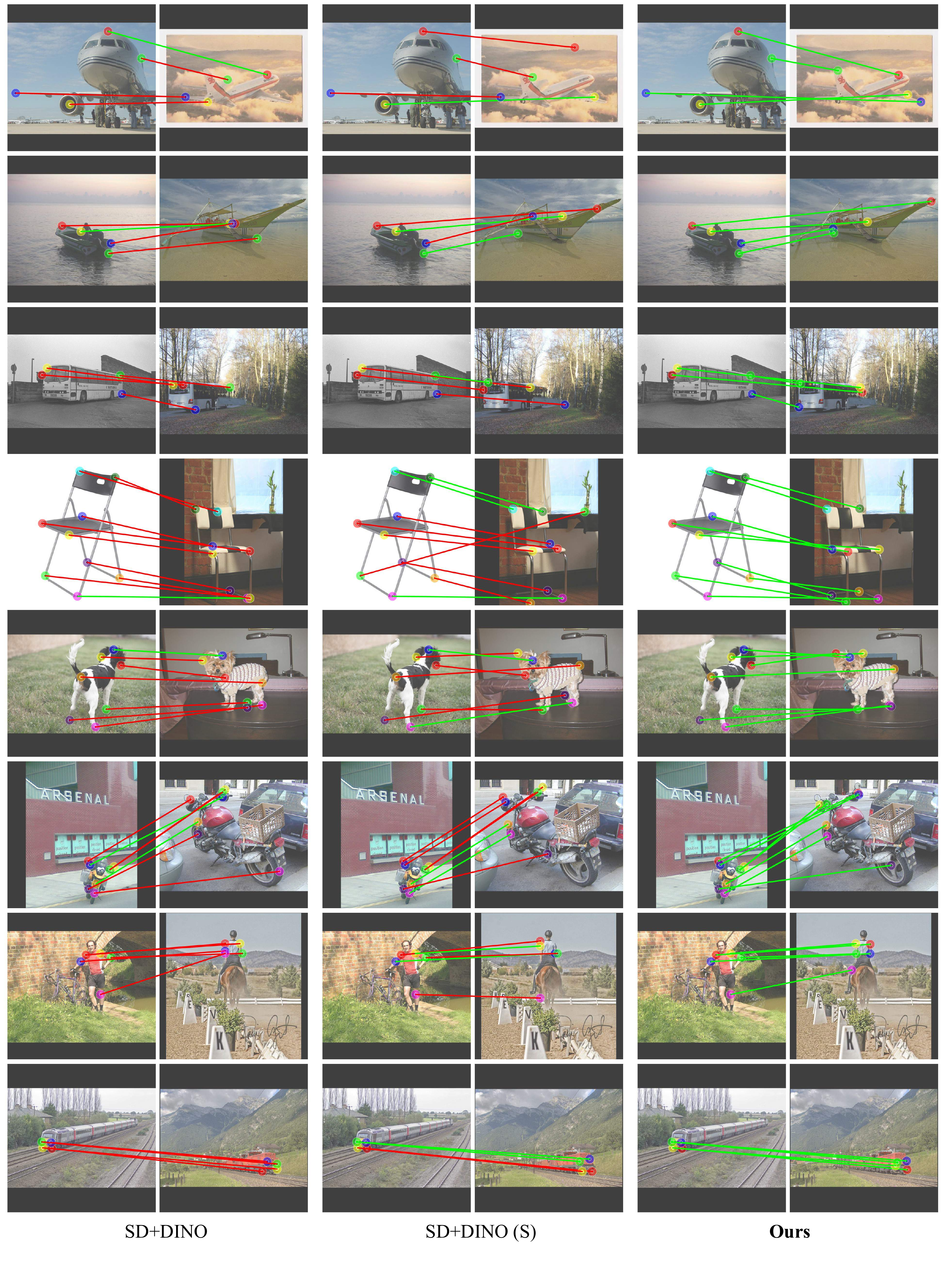}
   \caption{\textbf{Qualitative comparison on the SPair-71k.} Our method shines even in cases with large viewpoint variations.}
   \label{fig:8qual_results_spair1}
\end{figure*}

\begin{figure*}[t]
\centering
   \includegraphics[width=0.95\linewidth]{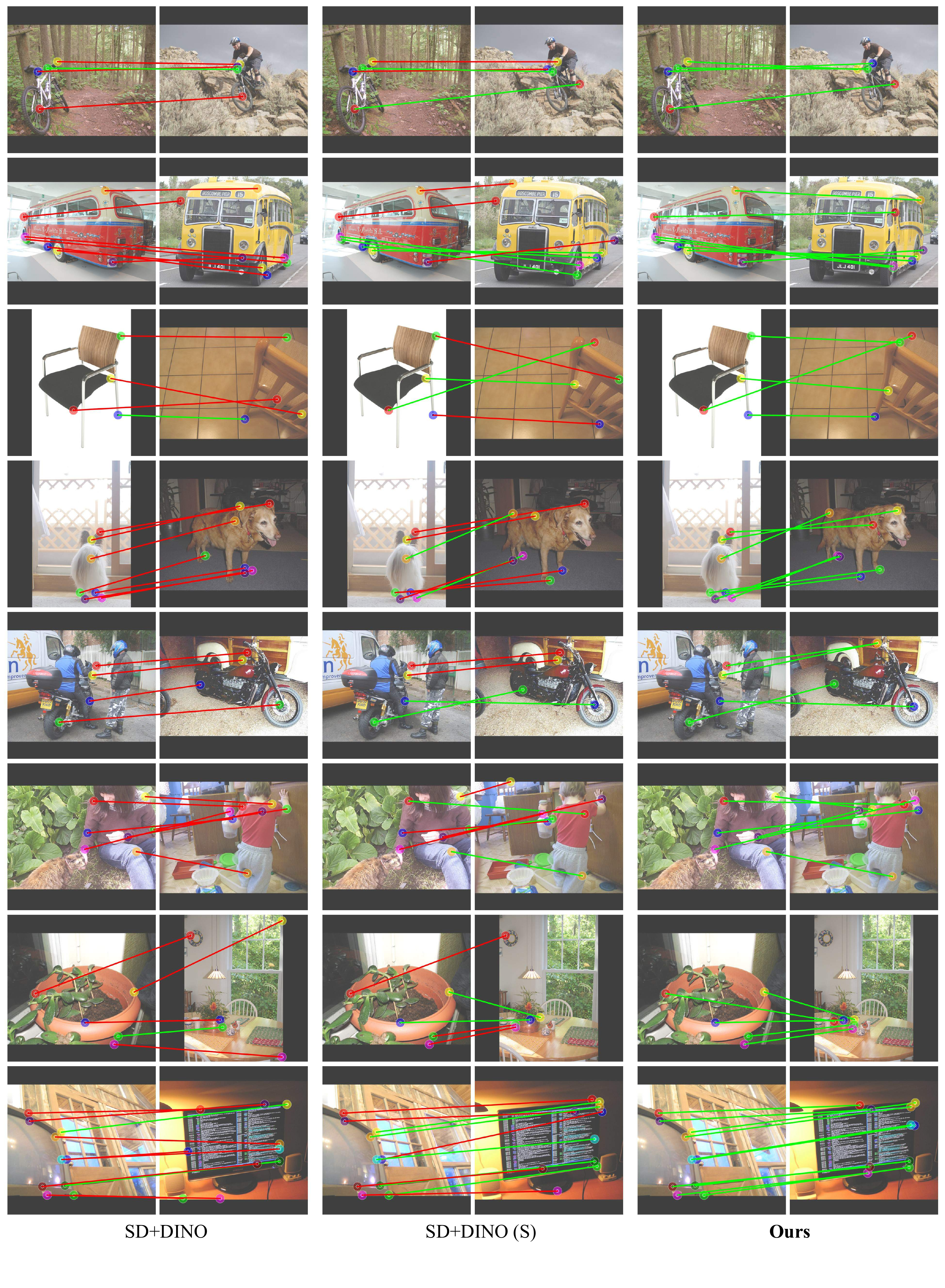}
   \caption{\textbf{Qualitative comparison on the SPair-71k.} Our method shines even in cases with large viewpoint variations.}
   \label{fig:8qual_results_spair2}
\end{figure*}

In~\cref{fig:8qual_results_spair1} and~\cref{fig:8qual_results_spair2}, we show the qualitative comparison of our supervised methods with both the unsupervised and supervised versions of SD+DINO~\cite{zhang2023tale} on SPair-71k dataset.
Our method establishes correct correspondence for challenging cases that previous works cannot handle.

\end{document}